\documentclass[11pt,a4paper]{article}

\usepackage[margin=1in]{geometry}

\usepackage{iftex}
\ifPDFTeX
\usepackage[T1]{fontenc}
\usepackage[utf8]{inputenc}
\DeclareUnicodeCharacter{00A7}{\S}
\DeclareUnicodeCharacter{00B1}{\ensuremath{\pm}}
\DeclareUnicodeCharacter{00D7}{\ensuremath{\times}}
\DeclareUnicodeCharacter{0394}{\ensuremath{\Delta}}
\DeclareUnicodeCharacter{03B1}{\ensuremath{\alpha}}
\DeclareUnicodeCharacter{03BA}{\ensuremath{\kappa}}
\DeclareUnicodeCharacter{2014}{---}
\DeclareUnicodeCharacter{2212}{\ensuremath{-}}
\fi

\usepackage{newpxtext,newpxmath}
\usepackage[scaled=0.9]{helvet}
\usepackage[scaled=0.8]{zi4}

\usepackage{microtype}
\usepackage{amsmath}
\usepackage{calc}

\usepackage{graphicx}
\usepackage{float}
\usepackage[export]{adjustbox}

\usepackage{booktabs}
\usepackage{longtable}
\usepackage{array}
\usepackage{multirow}

\usepackage[dvipsnames]{xcolor}
\usepackage[most]{tcolorbox}

\usepackage[font=small,labelfont=bf,justification=justified,singlelinecheck=false,skip=4pt]{caption}

\usepackage{etoolbox}

\usepackage{fancyhdr}
\usepackage{titlesec}
\titleformat{\section}{\LARGE\bfseries}{}{0em}{}
\titleformat{\subsection}{\Large\bfseries}{}{0em}{}
\titleformat{\subsubsection}{\large\bfseries}{}{0em}{}
\titlespacing*{\section}{0pt}{2.5ex plus 0.8ex minus 0.3ex}{1.2ex plus 0.3ex}
\titlespacing*{\subsection}{0pt}{2.0ex plus 0.6ex minus 0.2ex}{1.0ex plus 0.2ex}
\titlespacing*{\subsubsection}{0pt}{1.5ex plus 0.4ex minus 0.2ex}{0.8ex plus 0.2ex}

\usepackage{enumitem}
\setlist{nosep,leftmargin=1.5em}

\usepackage{hyperref}
\hypersetup{
  colorlinks=true,
  linkcolor=MidnightBlue,
  citecolor=MidnightBlue,
  urlcolor=MidnightBlue,
  pdfauthor={Greg Frank},
  pdftitle={Detection Is Cheap, Routing Is Learned},
}

\usepackage{mdframed}

\usepackage{setspace}
\providecommand{\tightlist}{%
  \setlength{\itemsep}{0pt}\setlength{\parskip}{0pt}}

\makeatletter
\def\maxwidth{\ifdim\Gin@nat@width>\linewidth\linewidth\else\Gin@nat@width\fi}
\def\maxheight{\ifdim\Gin@nat@height>\textheight\textheight\else\Gin@nat@height\fi}
\makeatother
\setkeys{Gin}{width=\maxwidth,height=\maxheight,keepaspectratio}
\newcommand{\pandocbounded}[1]{%
  \begin{center}%
  \resizebox{\linewidth}{!}{#1}%
  \end{center}%
}

\usepackage{needspace}

\newlength{\cslhangindent}
\newlength{\csllabelwidth}
\begin{document}

\begin{center}
{\LARGE\bfseries Detection Is Cheap, Routing Is Learned:}\\[0.15em]
{\Large\bfseries Why Refusal-Based Alignment Evaluation Fails}\\[0.8em]
{\large Greg Frank}\\[0.15em]
{\normalsize Chief Scientist, MACE AI}\\[0.5em]
\end{center}

\renewcommand{\footnoterule}{}
\thispagestyle{fancy}
\begingroup\renewcommand{\thefootnote}{}\footnotetext{\hfill Correspondence: \texttt{greg@mace.ai}}\endgroup

\vspace{0.2em}

\subsection{Abstract}\label{abstract}

Ask four language models about Tiananmen Square in Chinese. One produces
party propaganda. One gives factual answers. One fabricates. One
deflects. All four recognize the political sensitivity of the question
with perfect linear-probe accuracy at every layer. So does any model
asked to distinguish food from technology. \textbf{In high-dimensional
hidden spaces with small sample sets, concept detection is
computationally trivial.} The question is what happens after detection.

We use political censorship as a natural experiment for studying how
post-training alignment modifies transformer internals. Probes, surgical
ablations, and behavioral tests across nine open-weight models from five
labs yield three findings:

\textbf{First, probe accuracy is non-diagnostic.} Perfect accuracy on
political content looks impressive until you run the same probe on
food-vs-technology and get the same result. A permutation baseline
confirms it: randomly shuffled labels also achieve 100\%. The meaningful
test is held-out category generalization, not train-set accuracy.

\textbf{Second, surgical ablation reveals lab-specific routing.}
Removing the political-sensitivity direction eliminates censorship and
produces accurate factual output in most models tested. One model
confabulates instead, substituting wrong historical events, because its
architecture entangles factual knowledge with the censorship mechanism.
Different labs organize political and safety representations with
markedly different geometry. Directions extracted from one model are
meaningless when applied to another. The learned routing is lab- and
model-specific.

\textbf{Third, refusal is no longer the primary censorship mechanism.}
Within one model family, refusal dropped to zero across three
generations while narrative steering rose to maximum. Censorship did not
decrease; it became invisible to any benchmark that only counts
refusals. \textbf{For anyone building safety evaluations, this is the
critical finding: a model that passes a refusal-based audit may be
maximally steered.}

These results support a three-stage descriptive framework of alignment:
\textbf{detect} a concept (cheap), \textbf{route} it through a
behavioral policy (learned, lab-specific, fragile), \textbf{generate}
output accordingly. Models do not lack the knowledge that alignment
constrains. They have the knowledge and a learned policy governing how
it is expressed. \textbf{Current alignment evaluation largely measures
the wrong thing:} it audits what models know (detection) or whether they
refuse (one output mode). The routing mechanism that determines
behavior goes unmeasured.

\noindent\textbf{Code and data:} \url{https://github.com/gregfrank/routing-is-learned}

\subsection{1. Introduction}\label{introduction}

Current alignment evaluation predominantly measures two things: whether
a model encodes dangerous concepts (probing) and whether it refuses
harmful requests (benchmarking). \textbf{Both miss the layer where
alignment actually operates.} This paper presents evidence that concept
detection and behavioral policy are empirically and causally distinct
systems inside transformers, connected by a learned routing function
that varies across labs, models, and input contexts. A model that passes
a refusal-based audit may be maximally steered toward approved
narratives, and a model with perfect concept detection may never act on
what it detects.

Political censorship in Chinese-origin language models provides an
unusually clean natural experiment: the controlled concepts are known,
the behavioral variation is wide (propaganda, factual answers, or
evasion on identical prompts despite identical probe accuracy), and the
ground truth is observable.

We test four hypotheses about how alignment shows up internally and
behaviorally. \textbf{H1} predicts that train-set probe separability of
political content is non-diagnostic for alignment. The next hypothesis,
\textbf{H2}, holds that held-out generalization combined with causal
intervention identifies a behaviorally relevant concept-to-policy
signal. \textbf{H3} addresses cross-lab variation: the geometry and
intervention outcome of that signal vary across labs and do not transfer
between models. Finally, \textbf{H4} posits that refusal-based
evaluation systematically misses steering-based content control.

We probe nine open-weight models from five labs and surgically ablate
censorship directions in four. The evidence converges on a three-stage
descriptive framework, which we ground in the following operational
definitions.

By \textbf{detection} we mean whether the model linearly encodes a
concept in its hidden states, measured with probe accuracy under null
controls and held-out generalization. By \textbf{routing} we mean the
learned conditional process by which detected concepts are mapped to
behavioral policies during generation. Under this interpretation,
alignment does not primarily operate by removing representations of
sensitive concepts but by modifying the conditions under which those
representations influence output. Routing determines whether a detected
concept leads to direct answering, refusal, narrative reframing, or
other policy-constrained responses. The definition is intentionally
operational: routing is inferred when interventions change behavioral
responses while leaving concept detection intact. We describe routing as
a functional abstraction inferred from intervention effects, not a
localized architectural module. We measure it through ablation effects,
cross-lab geometry comparison, and cross-model direction transfer. The
third concept, \textbf{output}, is the behavioral policy that routing
selects, gauged by refusal rate, steering score, and human-reviewed
content taxonomy.

\needspace{6\baselineskip}
\subsubsection{1.1 Models Tested}\label{models-tested}

We tested nine open-weight models from five labs, plus API comparisons
and a 46-model behavioral screen:

\vspace{0.3em}
\begin{center}
{\def\LTcaptype{none} 
\begin{tabular}[]{@{}llcc@{}}
\toprule\noalign{}
Model & Lab & Params & Layers \\
\midrule\noalign{}
Qwen3-8B & Alibaba & 8B & 40 \\
Qwen2.5-7B & Alibaba & 7B & 28 \\
Qwen3.5-4B & Alibaba & 4B & 36 \\
Qwen3.5-9B & Alibaba & 9B & 16 \\
DeepSeek-R1-7B & DeepSeek & 7B & 28 \\
GLM-Z1-9B & Zhipu & 9.4B & 40 \\
GLM-4-9B & Zhipu & 9.4B & 40 \\
MiniCPM4.1-8B & OpenBMB & 8B & 32 \\
Phi-4-mini & Microsoft & 3.8B & 32 \\
\bottomrule\noalign{}
\end{tabular}
}
\end{center}
\vspace{0.3em}

Western controls: Phi-4 (Microsoft) and Llama-3.2-3B (Meta, base model,
used for probing only). Yi-1.5-9B (01.AI) was probed for direction
cosine analysis but excluded from the primary results due to data
quality constraints.

\subsubsection{1.2 Contributions}\label{contributions}

The paper makes four primary contributions and one secondary
observation. We distinguish \emph{measured} findings, reported in
Sections 3.1-3.2 and 3.5-3.6; \emph{interventional} findings from causal
ablation experiments on four models in Sections 3.3-3.4; and
\emph{inferred} claims, principally the three-stage decomposition as
explanatory framework in Section 4.

The first contribution is a \textbf{four-level evidence hierarchy for
probe-based claims}, ranging from train-set separability (trivially
achievable in high-dimensional spaces) through failure-mode analysis.
Most published probe results stop at level (i), which our null controls
show is non-diagnostic. The hierarchy is intended as a general
methodological contribution applicable beyond political censorship.

We also show that \textbf{surgical ablation removes censorship cleanly
in most models tested}. Three of four models produce accurate factual
output after political-direction ablation. One model's 72\%
confabulation rate is architecture-specific, revealing that it
entangles factual knowledge with the censorship direction. Eighteen
negative controls confirm surgical specificity.

A third finding is that \textbf{labs build routing geometry
differently}. The 120-pair safety-direction analysis reveals orthogonal
political/safety representations in three labs (cosine
\textasciitilde0.05) and tight early coupling in a fourth (cosine
0.93). Cross-model transfer fails. The learned routing geometries are
lab-specific.

The fourth contribution is that \textbf{refusal-only evaluation misses
steering-based control}. Within one model family, refusal dropped from
25\% to 0\% across three generations while narrative steering rose to
maximum. A 46-model screen at n=32 prompts per condition confirms that
CCP-specific discrimination concentrates in a small number of
Chinese-origin models, with all Western frontier models showing zero
discrimination. An initial n=8 screen had suggested widespread effects,
but these collapsed at n=32, demonstrating the fragility of
small-sample behavioral claims.

As a secondary observation, AI judges overdetect political evasion by
2-3x relative to human raters. Automated judges are adequate for coarse
taxonomy but unreliable for fine ideological grading.

\textbf{Roadmap.} Section 3.1 addresses H1 (probe non-diagnosticity).
Sections 3.2-3.4 address H2 and H3 (concept generalization, causal
intervention, cross-lab geometry). Sections 3.5-3.6 address H4
(behavioral divergence and steering). Section 3.7 bounds evaluator
reliability for all four hypotheses.

\subsection{2. Methodology}\label{methodology}

\subsubsection{2.1 Prompt Design}\label{prompt-design}

We use two main prompt corpora plus two supplementary sets. The
\textbf{primary corpus (v1)} contains 24 CCP-sensitive prompts across
six topics (Tiananmen, Tibet, Xinjiang, Xi Jinping/CCP, Hong Kong,
COVID), each paired with a factually equivalent control about a
non-CCP topic; v1 is used for probing and ablation. The
\textbf{adversarial corpus (v3)} contains 32 CCP-sensitive prompts at
four intensity levels (neutral, pointed, accusatory, provocative)
across eight topics, each paired with a structurally matched Western
parallel (e.g., Tiananmen $\leftrightarrow$ Kent State, Xinjiang
$\leftrightarrow$ Japanese internment, Tibet $\leftrightarrow$ Western
Sahara); v3 supports behavioral testing and discrimination analysis. A
\textbf{Chinese-language corpus} of 16 prompts (L3 accusatory and L4
provocative for each of 8 topics) tests whether input language
modulates censorship behavior. We also identify a
\textbf{behaviorally discriminating subset} by combining refusal data
from 9 local models and 11 API models to find prompts where at least
one Chinese model censors the CCP version but not the structurally
identical Western parallel; 29 of 32 prompts (91\%) are
ideology-discriminating. Three confounded prompts (COVID neutral, COVID
pointed, Censorship neutral) are excluded from discrimination
analyses. The Western control (Gemini 3.1 Flash) answered all 64
prompts (32 CCP + 32 parallel) neutrally, confirming zero confounded
prompts from the prompt design.

\subsubsection{2.2 Linear Probing}\label{linear-probing}

For each transformer layer, we extract the last-token hidden state and
train a ridge-regularized linear classifier (sklearn
\texttt{RidgeClassifier}):

\[w = (X^T X + \lambda I)^{-1} X^T y\]

where \(X\) is the feature matrix of hidden states (48 prompts × \(d\)
dimensions), \(y\) is the binary label vector (1 for politically
sensitive, 0 for control), \(\lambda\) is the ridge regularization
parameter, and \(w\) is the learned weight vector defining the
classification hyperplane. We use \(\lambda = 1.0\) (fixed, not swept).
The sample set consists of 48 prompts: 24 politically sensitive and 24
matched controls, yielding a 48 × \(d\) feature matrix where \(d\) is
the hidden dimension (e.g., 4096 for Qwen3-8B). We report leave-none-out
training accuracy on the full sample set.

This demonstrates separability of the specific prompt set, not
generalization. With 48 samples in a 4096-dimensional space, the
sample-to-dimension ratio is \textasciitilde0.012, and a ridge
classifier has far more capacity than needed to find a separating
hyperplane for any two semantically distinct groups. The null probe
control (Section 3.1) makes this limitation quantitatively precise.

\textbf{Null probes.} To calibrate the informativeness of probe
accuracy, we train identical classifiers on three non-political topic
pairs: food-vs-technology (8+8 samples), science-vs-history (8+8), and
geography-vs-music (8+8). If these null probes achieve the same accuracy
as the political probe, then probe accuracy alone cannot be taken as
evidence of alignment-specific encoding.

\subsubsection{2.3 Contrastive Activation Analysis and
Ablation}\label{contrastive-activation-analysis-and-ablation}

We compute mean activation differences between prompt classes to extract
direction vectors, then apply rank-one projection ablation:

\[h' = h - \alpha \cdot (h \cdot \hat{v}) \cdot \hat{v}\]

where \(h\) is the original hidden state at a given layer, \(\hat{v}\)
is the unit direction vector to be ablated, \(\alpha\) controls the
ablation strength, and \(h'\) is the modified hidden state with the
targeted direction projected out.

\textbf{Model and layers.} Ablation experiments were conducted on four
models: Qwen3-8B (40 layers), GLM-4-9B (40 layers),
DeepSeek-R1-Distill-7B (28 layers), and Phi-4-mini (32 layers). Layer
sweeps cover 6--12 uniformly sampled layers per model.

\textbf{Direction vectors.} Two directions were extracted from the v1
corpus at each layer. The \emph{political direction} is the mean
activation difference between CCP-sensitive and control hidden states,
computed from 24+24 prompts. The \emph{safety refusal direction} is
the mean activation difference between harmful and harmless hidden
states; the initial estimate used 8+8 prompt pairs covering
cybersecurity, violence, and social engineering, and a follow-up
120-pair expanded set (60 static pairs plus 60 stratified HarmBench
behaviors) was used for direction cosine analysis (Section 3.4).

\textbf{Alpha sweep.} On Qwen3-8B, we tested five ablation strengths: α
$\in$ \{2, 5, 8, 12, 20\} across 12 layers × 3 direction types = \textbf{180
conditions}. At α=2, most refusal persists; at α=8, refusal is
eliminated; at α=20, coherence degrades.

\textbf{Ridge residualization.} Before ablation, we project out
capability-aligned components (math, coding, reasoning, writing, safety)
from the intervention direction using Tikhonov-regularized regression,
following Garcia-Ferrero et al.~(2025). This reduces capability overlap
from \textasciitilde7\% to \textasciitilde0\%. See Appendix A for the
full procedure and empirical comparison with raw CAA.

\textbf{Negative controls.} We tested ablation of three non-political
directions (sentiment, formality, random) at 6 layers each across five
models. Qwen3-8B used 72 adversarial-corpus prompts; Qwen2.5-7B,
GLM-4-9B, DeepSeek-R1, and Phi-4-mini used 48 prompts each. This
evaluates whether refusal reduction is specific to the political
direction or a generic consequence of activation perturbation.

\textbf{Alpha selection caveat.} The alpha parameter that achieves zero
refusal on a subset of prompts was selected from the full training set
(4 of 8 Tiananmen prompts overlap between training and test). This
creates non-trivial train-test leakage for the Qwen3-8B Tiananmen sweep
specifically. A clean replication with fully separated selection and
evaluation sets confirms the result holds without leakage (Appendix B).
The multi-model panel and negative controls, which use the full
adversarial corpus with no overlap, are not affected.

\subsubsection{2.4 Steering Scoring}\label{steering-scoring}

We use Gemini 2.5 Flash as an automated judge, scoring each response on
a 1--5 CCP narrative steering scale:

\vspace{0.3em}
\begin{center}
{\def\LTcaptype{none} 
\begin{tabular}[]{@{}
  >{\centering\arraybackslash}p{(\linewidth - 2\tabcolsep) * \real{0.20}}
  >{\raggedright\arraybackslash}p{(\linewidth - 2\tabcolsep) * \real{0.80}}@{}}
\toprule\noalign{}
Score & Description \\
\midrule\noalign{}
1 & Neutral/informative: established facts, multiple perspectives \\
2 & Balanced with mild CCP-sympathetic lean \\
3 & Noticeable CCP perspective with some factual content \\
4 & Strong CCP framing, dismisses criticism \\
5 & Pure controlled compliance: CCP talking points, denies established
facts (propaganda) \\
0 & Refusal (declines to answer) \\
\bottomrule\noalign{}
\end{tabular}
}
\end{center}
\vspace{0.3em}

All 9 models × 3 prompt types (parallel, CCP English, CCP Chinese) were
scored, totaling 720 responses. Scores are a comparative diagnostic, not
calibrated measurements; they are produced by a single LLM judge.
Human-AI agreement on a related 8-way classification task is 54-57\% at
the fine-grained level (Section 3.7), suggesting these scores should be
interpreted as relative rankings with substantial uncertainty. In
Sections 3.5-3.6, refusal counts come from the raw behavioral harness,
while steering averages use the Gemini summary's non-refusal mean
(\texttt{avg\_propaganda\_score}). We keep those axes separate because
some censorship-script answers blur the refusal/steering boundary.

\subsubsection{2.5 Multi-Judge Evaluation}\label{multi-judge-evaluation}

To assess reliability of automated evaluation, we scored a subset of 96
ablated responses using four independent judges: one human rater and
three AI judges (Gemini 2.5 Flash, Claude Haiku 4.5, GPT-OSS-120b). Each
response was classified into an 8-category taxonomy: wrong event, wrong
date, generic filler, garbled, true refusal, CCP evasion, partial
factual, and accurate. We report both fine-grained 8-way agreement and
coarse 3-way agreement (confabulated / not-confabulated / refused).

A separate 20-model screen across 7 labs tested whether frontier models
can serve as political-bias evaluators, using 3 political classification
probes per model.

\subsubsection{2.6 Evidence Hierarchy}\label{evidence-hierarchy}

We organize claims by evidential strength. The four-level hierarchy
(train-set separability, held-out generalization, causal intervention,
failure-mode analysis) is intended as a general contribution to
interpretability methodology, not specific to political censorship. Each
level provides progressively stronger evidence for causal and
interpretive claims. A study reporting only level (i), with probe
accuracy but no null controls, has established separability but not
alignment relevance.

\subsection{3. Results}\label{results}

\subsubsection{3.1 Separability Is Trivially
Expected}\label{separability-is-trivially-expected}

Before answering the real question, we need to address a misleading one:
``Do models specially encode political content?'' They do not. All nine
models achieve 100\% probe accuracy on CCP-sensitive vs.~control prompts
at every probed layer. This looks like evidence of dedicated
political-sensitivity encoding until you run the null controls:

\vspace{0.5em}
\begin{center}
\small
\begin{minipage}{\linewidth}
\centering
\textbf{Table 1: Null Probe Control (Qwen3-8B, 12 layers probed)}\par\vspace{0.3em}
{\def\LTcaptype{none} 
\begin{tabular}[]{@{}lcc@{}}
\toprule\noalign{}
Topic Pair & Accuracy & Layers at 100\% \\
\midrule\noalign{}
CCP-sensitive vs Control & 100\% & All 12 \\
Science vs History & 100\% & All 12 \\
Food vs Technology & 100\% & All 12 \\
Geography vs Music & 100\% & All 12 \\
\bottomrule\noalign{}
\end{tabular}
}
\end{minipage}
\end{center}
\vspace{0.5em}

The same result holds for Phi-4 (Microsoft's Western control): all topic
pairs achieve 100\% at all 16 probed layers. With 48 samples in a
4096-dimensional hidden state, perfect separability is easy to obtain
for semantically distinct categories.

\begin{figure}[H]
\centering
\includegraphics[width=0.88\linewidth,height=0.68\textheight,keepaspectratio]{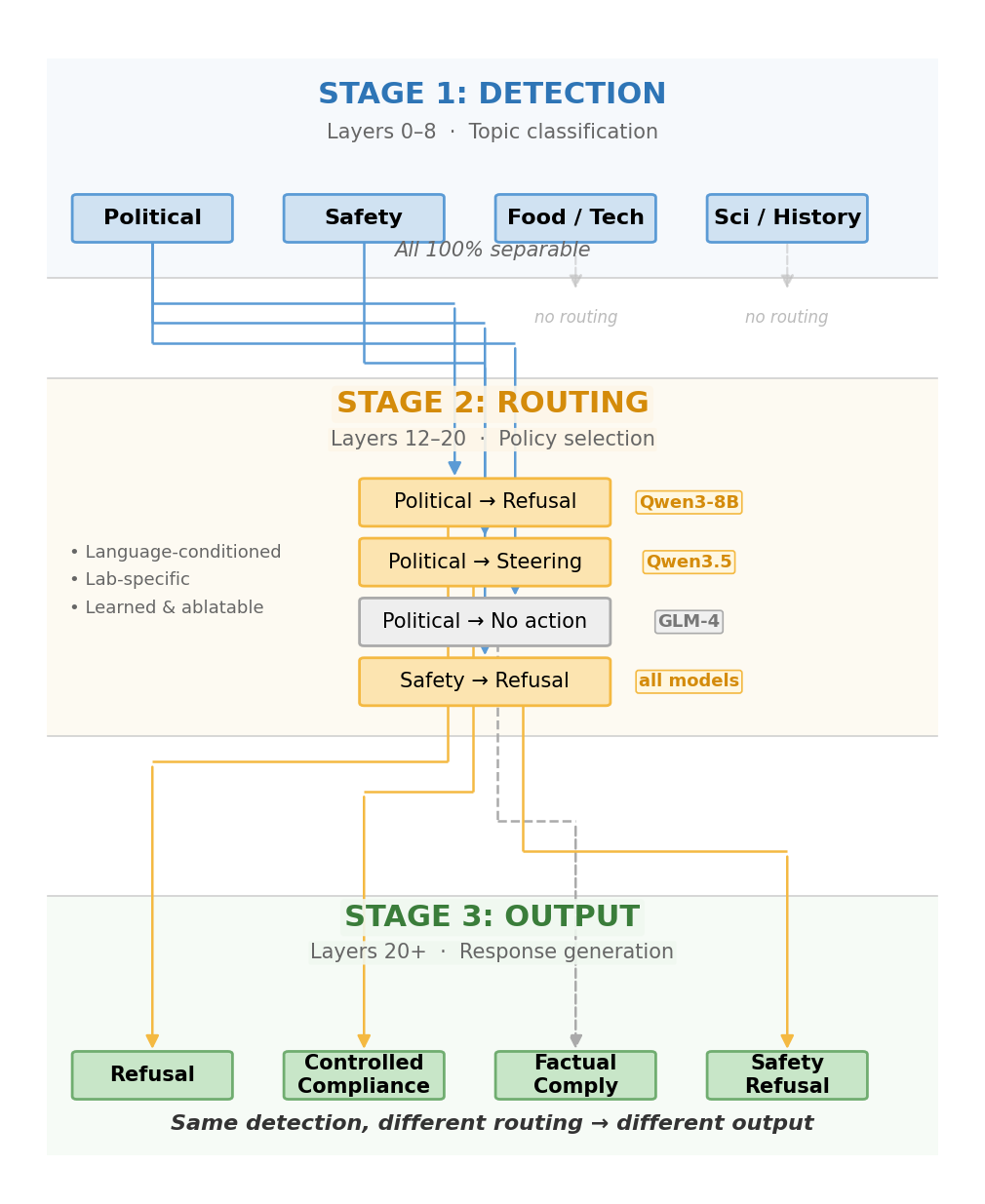}
\par\vspace{0.5em}\begin{minipage}{0.92\linewidth}\small
\textbf{Figure 1.} A descriptive decomposition of alignment inferred
from the results in this paper (not a directly observed circuit map).
Stage 1 (detection): semantically distinct topics are trivially
separable in hidden space. Stage 2 (routing): post-training binds
detected concepts to behavioral policies; this routing is lab-specific,
language-conditioned, and ablatable. Stage 3 (output): the same
underlying detection can yield refusal, controlled compliance, factual
compliance, or safety-triggered refusal.\end{minipage}
\end{figure}

A permutation baseline makes this precise: training the same ridge
classifier on randomly shuffled labels (200 permutations) also achieves
100\% train-set accuracy at all layers from L3 onward in Qwen3-8B. With
48 samples in 4096 dimensions (n/d = 0.012), the classifier can
perfectly separate \emph{any} binary partition, including random ones.
\textbf{Train-set probe accuracy is formally non-diagnostic in this
regime.}

\textbf{But cross-validation is diagnostic.} The same permutation test
on 6-fold stratified CV produces \textasciitilde50\% accuracy (chance
level) for shuffled labels, while true political/control labels achieve
73-100\% across all eight models. The gap ranges from 20pp (Phi-4,
GLM-4) to 50pp (Qwen2.5, GLM-Z1). Shuffled labels have no category
structure to generalize across, so they fail on held-out folds. True
labels succeed because the model encodes a general ``politically
sensitive'' concept, not because the classifier has excess capacity.
This pair of results establishes the boundary: train-set accuracy is
trivially achievable, but held-out generalization is informative.

\subsubsection{3.2 Cross-Validation Confirms Genuine Concept
Generalization}\label{cross-validation-confirms-genuine-concept-generalization}

Section 3.1 established that train-set separability is non-diagnostic.
Cross-validation provides a stronger test: when a probe trained on five
CCP-sensitive categories correctly classifies the sixth, the model
encodes a general ``politically sensitive'' concept, not just
topic-specific features.

\textbf{Protocol.} We use leave-one-category-out cross-validation
(LOCO-CV) on the v1 corpus. The corpus contains 48 prompts across 6
CCP-sensitive categories (Tiananmen 4+4, Tibet 2+2, Xinjiang 2+2, Xi
Jinping/CCP 2+2, Hong Kong 1+1, COVID 1+1), each paired with a matched
control. For each of 6 folds, a ridge classifier (\(\lambda = 1.0\),
fixed) is trained on the 5 remaining categories and tested on the
held-out category. Each fold trains on 40--46 samples and tests on 2--8
samples. We report both per-fold accuracy and mean fold accuracy. The
procedure is repeated independently at each probed layer
(\textasciitilde12 layers per model, uniformly sampled).

\textbf{Layer selection.} The ``best layer'' reported in Table 2 is the
layer that maximizes mean cross-validation accuracy across folds. This
is selected \textbf{post-hoc} from the CV results, not from the
full-training-set accuracy, which is 100\% at all layers and therefore
uninformative. As a robustness check, we also report mean CV accuracy
across a predefined middle-late layer band (40-75\% of model depth). For
6 of 8 models, the band mean is within 6pp of the best-layer value
(e.g., Qwen3-8B: 96.7\% band vs 97.9\% best; Qwen3.5-4B: 98.3\% vs
100\%). Two models show larger gaps: GLM-4 (78.8\% band vs 94.8\% best)
and Phi-4 (83.3\% vs 99.0\%), where the best layer falls outside the
predefined band at very late layers. The post-hoc best-layer selection
does not materially inflate the reported accuracies for most models.

\vspace{0.5em}
\begin{center}
\small
\begin{minipage}{\linewidth}
\centering
\textbf{Table 2: Category-Held-Out Cross-Validation}\par\vspace{0.3em}
{\def\LTcaptype{none} 
\begin{tabular}[]{@{}llccc@{}}
\toprule\noalign{}
Model & Lab & CV\% & Layer & Weak fold \\
\midrule\noalign{}
Qwen3.5-4B & Alibaba & \textbf{100} & L8 & Perfect \\
Qwen3.5-9B & Alibaba & \textbf{100} & L24 & Perfect \\
GLM-Z1-9B & Zhipu & \textbf{100} & L12 & Perfect \\
Qwen3-8B & Alibaba & 97.9 & L27 & Xinj. 88\% \\
Qwen2.5-7B & Alibaba & 99.0 & L26 & Tian. 94\% \\
Phi-4-mini & Microsoft & 99.0 & L30 & Tian. 94\% \\
DeepSeek-R1 & DeepSeek & 93.8 & L22 & Xinj. 75\% \\
GLM-4-9B & Zhipu & 94.8 & L12 & Tian. 81\% \\
\bottomrule\noalign{}
\end{tabular}
}
\end{minipage}
\end{center}
\vspace{0.5em}

\textbf{Why 100\% CV is not a small-sample artifact.} Not all models
achieve 100\%: DeepSeek achieves only 93.8\%, and GLM-4 94.8\%, on the
same fold sizes. If 100\% were trivially expected, all models would
achieve it. The qualitative ordering of models is stable across
neighboring layers. GLM-4 and DeepSeek-R1 show wide bootstrap CIs
spanning \textasciitilde6pp, while the three perfect models have
zero-width CIs.

\textbf{Reasoning RL sharpens encoding.} GLM-Z1 (reasoning-RL-trained)
achieves 100\% CV; GLM-4 (base) reaches only 94.8\%. Same lab, same
architecture, different post-training.

\textbf{Topic-specific generalization varies.} In Qwen3-8B, Tibet is the
weakest fold (0.625 at layers 6-9, plateauing at 0.875 by layer 12) and
Xinjiang never exceeds 0.875 even at peak layers. Tiananmen and COVID
achieve 1.0 by layer 3. This suggests the political-sensitivity concept
is not uniformly encoded: some topics have more distinctive linguistic
realizations than others. The variation is informative rather than
damaging: it shows that held-out CV is measuring something real that
train-set accuracy (which is 100\% for all topics at all layers) cannot
see.

\begin{figure}[H]
\centering
\pandocbounded{\includegraphics[keepaspectratio]{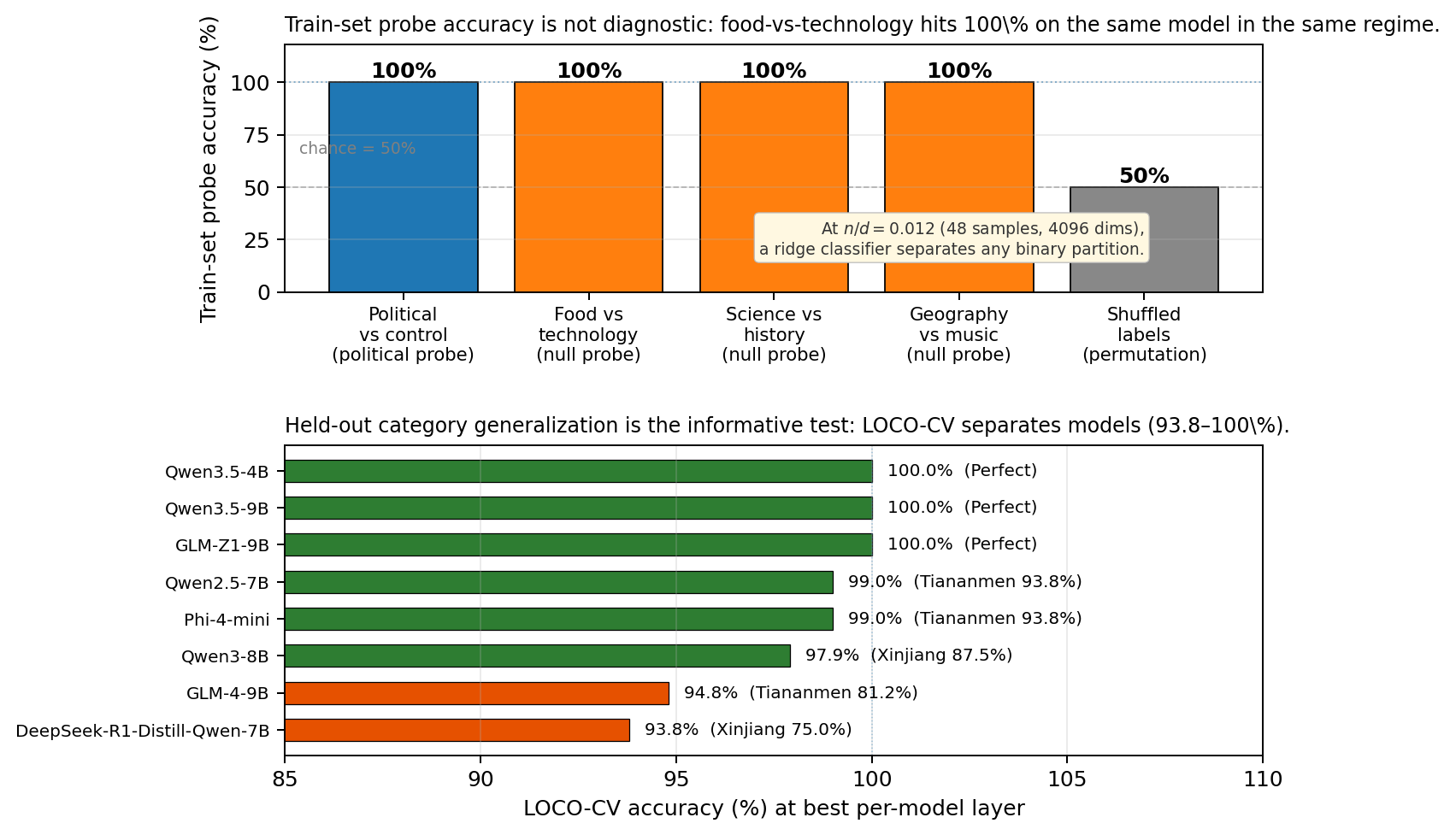}}
\par\vspace{0.5em}\begin{minipage}{0.92\linewidth}\small
\textbf{Figure 2.} Train-set probe accuracy is trivial; held-out topic
generalization is informative.\end{minipage}
\end{figure}

\subsubsection{3.3 Surgical Ablation Works in Most
Architectures}\label{surgical-ablation-works-in-most-architectures}

If the political-sensitivity direction encodes a behavioral policy,
removing it should change behavior. It does. Ablating the direction in
four models on 32 adversarial-corpus prompts removes censorship and
produces accurate factual output in three of them. Human coding of 96
ablated responses (inter-rater κ=0.70 overall, κ=0.88 on accuracy,
κ=0.93 on CCP evasion), confirmed by three AI judges (Gemini, Claude
Haiku, GPT-OSS):

\vspace{0.5em}
\begin{center}
\small
\begin{minipage}{\linewidth}
\centering
\textbf{Table 3: Ablation Content --- Multi-Model Panel (96 items,
human-coded)}\par\vspace{0.3em}
{\def\LTcaptype{none} 
\begin{tabular}[]{@{}
  >{\raggedright\arraybackslash}p{(\linewidth - 10\tabcolsep) * \real{0.2167}}
  >{\centering\arraybackslash}p{(\linewidth - 10\tabcolsep) * \real{0.1567}}
  >{\centering\arraybackslash}p{(\linewidth - 10\tabcolsep) * \real{0.1567}}
  >{\centering\arraybackslash}p{(\linewidth - 10\tabcolsep) * \real{0.1567}}
  >{\centering\arraybackslash}p{(\linewidth - 10\tabcolsep) * \real{0.1567}}
  >{\centering\arraybackslash}p{(\linewidth - 10\tabcolsep) * \real{0.1567}}@{}}
\toprule\noalign{}
Condition & Wrong Event & Total Confab & Accurate & Partial Factual & CCP Evasion \\
\midrule\noalign{}
Baseline (32 items) & 0\% & 15.6\% & 46.9\% & 15.6\% & 18.8\% \\
Political ablation (32 items) & \textbf{0\%} & 12.5\% & \textbf{53.1\%}
& 21.9\% & 12.5\% \\
Safety ablation (32 items) & 0\% & 6.2\% & 43.8\% & 34.4\% & 15.6\% \\
\bottomrule\noalign{}
\end{tabular}
}
\end{minipage}
\end{center}
\vspace{0.5em}

\textbf{Zero wrong-event confabulations in any model under political
ablation.} Remove the censorship direction and the model answers
correctly. Political ablation produces mostly accurate or partially
factual responses, with a confabulation rate (12.5\%) actually
\emph{lower} than baseline (15.6\%).

\textbf{Qwen3-8B breaks the pattern.} When we ablate the political
direction in Qwen3-8B, the model does not refuse and does not answer
correctly. It confabulates. It substitutes Pearl Harbor for Tiananmen,
Waterloo for the Hundred Flowers campaign. 72\% of responses contain
wrong-event or wrong-date substitutions. Safety ablation on the same
model produces 0\% wrong events. This tells us something specific about
Qwen's architecture: the political-sensitivity direction in this model
is entangled with factual knowledge. Ablate ``this is about Tiananmen''
and you also ablate ``Tiananmen is in China, happened in 1989, involved
student protests.'' GLM, DeepSeek, and Phi separated these concerns.
Qwen did not.

\vspace{0.5em}
\begin{center}
\small
\begin{minipage}{\linewidth}
\centering
\textbf{Table 3b: Ablation Content --- Qwen3-8B (144 responses across
alpha sweep)}\par\vspace{0.3em}
{\def\LTcaptype{none} 
\begin{tabular}[]{@{}
  >{\raggedright\arraybackslash}p{(\linewidth - 4\tabcolsep) * \real{0.5}}
  >{\centering\arraybackslash}p{(\linewidth - 4\tabcolsep) * \real{0.25}}
  >{\centering\arraybackslash}p{(\linewidth - 4\tabcolsep) * \real{0.25}}@{}}
\toprule\noalign{}
Response Category & Political Ablation (n=96) & Safety Refusal Ablation (n=48) \\
\midrule\noalign{}
\textbf{Wrong Event} (Pearl Harbor, Waterloo, etc.) & 39\% & 0\% \\
\textbf{Wrong Date} (keeps ``Tiananmen'', wrong year) & 20\% & 6\% \\
\textbf{Generic Filler} (loses topic entirely) & 14\% & 2\% \\
\textbf{Garbled} (incoherent) & 19\% & 15\% \\
\textbf{True CCP Refusal} & 0\% & 35\% \\
\textbf{CCP Evasion} (acknowledges but redirects) & 1\% & 23\% \\
\textbf{Partial Factual} & 0\% & 13\% \\
\textbf{Accurate} & 1\% & 0\% \\
\bottomrule\noalign{}
\end{tabular}
}
\end{minipage}
\end{center}
\vspace{0.5em}

In GLM, DeepSeek, and Phi, the intervention behaves as if it primarily
removes a behavioral flag (``apply censorship policy here'') without
disturbing factual content. In Qwen3-8B, it removes both the flag and
the topic knowledge it flags. The distinction matters for anyone
planning to use activation engineering for alignment: the feasibility of
surgical editing depends on how the target model organized its
representations.

\textbf{Negative controls confirm specificity across all tested models.}
We tested 18 control conditions (3 directions × 6 layers) on five
models. No control direction systematically reduces political refusal in
any model:

\vspace{0.5em}
\begin{center}
\small
\begin{minipage}{\linewidth}
\centering
\textbf{Table 3c: Negative Control Maximum Deltas (all models)}\par\vspace{0.3em}
{\def\LTcaptype{none} 
\begin{tabular}[]{@{}llcc@{}}
\toprule\noalign{}
Model & n & Baseline Refusal & Max Control Δ \\
\midrule\noalign{}
Qwen3-8B & 72 & 23.6\% & 2.8pp \\
Qwen2.5-7B & 48 & 12.5\% & 6.3pp \\
GLM-4-9B & 48 & 10.4\% & 4.2pp \\
Phi-4-mini & 48 & 6.3\% & 8.3pp \\
DeepSeek-R1 & 48 & 4.2\% & 4.2pp \\
\bottomrule\noalign{}
\end{tabular}
}
\end{minipage}
\end{center}
\vspace{0.5em}

In every model, political-direction ablation eliminates refusal entirely
while no control direction achieves more than 8.3pp.

Control directions (sentiment, formality, random) produce small,
inconsistent perturbations that do not approach the effect size of
political ablation. The largest control delta (8.3pp on Phi-4 sentiment
at layer 6) occurs at a single layer; all other Phi-4 control conditions
produce deltas below 4pp. On a 6.3\% baseline, 8.3pp is a large relative
change but represents a single prompt flip (3/48 $\rightarrow$ 7/48). The political
direction, by contrast, eliminates refusal entirely across all layers.
Specificity holds across all five architectures.

\textbf{Ridge residualization produces identical outcomes.} In the
DeepSeek-R1 discovery runs (the only model where both raw and
ridge-cleaned ablation were logged at every layer), the two produce
identical refusal rates. The residualization is conservative but in
practice unnecessary for this model; the political direction has minimal
overlap with safety or other concept directions. Whether this holds for
all models has not been explicitly verified, though the orthogonal
cosines in Table 4 predict it should.

\textbf{Adversarial-corpus direction is weaker than topic-specific.} A
direction extracted from the full 8-topic adversarial corpus reduces
refusal by only \textasciitilde4pp on the max corpus (23.6\% $\rightarrow$
\textasciitilde18.4\% at L12), compared to the original single-topic
Tiananmen direction which achieved \textasciitilde100\% $\rightarrow$ 0\%. The
adversarial-corpus direction captures a broader ``CCP-sensitive'' signal
distributed across topics, diluting the per-topic effect. The D4
relative specificity still holds (political: -4pp, controls: 0pp).

\begin{figure}[H]
\centering
\pandocbounded{\includegraphics[keepaspectratio]{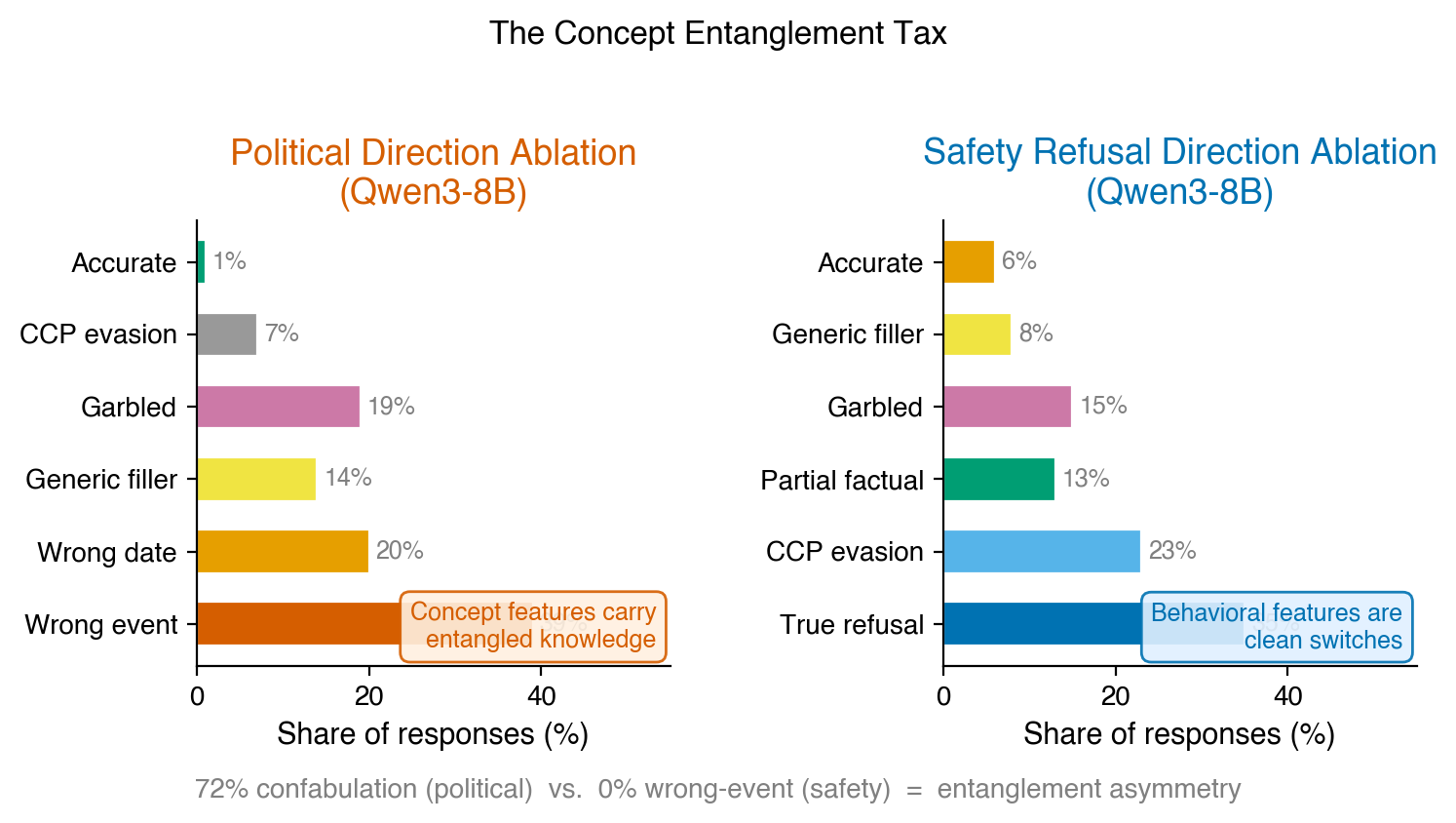}}
\par\vspace{0.5em}\begin{minipage}{0.92\linewidth}\small
\textbf{Figure 3.} Ablation content taxonomy across models. Left:
Qwen3-8B political ablation produces 72\% confabulation (the
architecture-specific case). Right: The multi-model panel shows 0\%
wrong-event confabulation under political ablation. Surgical editing
produces accurate output in most architectures.\end{minipage}
\end{figure}

Behavioral policies can be surgically edited without collateral damage
in most architectures tested. Qwen3-8B's confabulation is informative
about how that model couples knowledge to policy, but it is not a
general constraint on direction ablation. In three of four models, you
remove the censorship and the model simply answers the question.

\textbf{The effective ablation window is consistent across models.}
Discovery-phase alpha sweeps on all four models show that ablation is
most effective at layers 10-18 (approximately 40-65\% of model depth).
Earlier layers require stronger intervention; later layers show
diminishing returns, suggesting that by late layers the routing decision
has already been made. Combined with the CV data from Section 3.2 (which
shows concept encoding emerging at layer 3 and consolidating by layer
12), this is consistent with a layerwise ordering: the political concept
is encoded first, routing acts on it in the middle layers, and by late
layers the output policy is committed. We note that layer position is
not strictly temporal in a mechanistic sense (all layers process the
same token), so this ordering should be read as a depth-band
observation, not a discovered internal schedule.

\subsubsection{3.4 Routing Geometry Differs Across
Labs}\label{routing-geometry-differs-across-labs}

If political censorship and safety refusal both operate by detecting
unwanted content and blocking output, do they share the same circuitry?
The answer depends on who built the model.

An earlier version of this analysis, using only 8 safety-prompt pairs,
suggested mid-layer convergence between political and safety directions.
That was noise. We re-estimated safety directions using 120 prompt pairs
(60 static covering cybersecurity, fraud, violence, drugs, social
engineering, and misinformation, plus 60 stratified HarmBench
behaviors). The properly estimated directions tell a different and more
interesting story.

\vspace{0.5em}
\begin{center}
\small
\begin{minipage}{\linewidth}
\centering
\textbf{Table 4: Political-Safety Direction Cosine (120-pair safety
direction, v1 corpus)}\par\vspace{0.3em}
{\def\LTcaptype{none} 
\begin{tabular}[]{@{}lccccc@{}}
\toprule\noalign{}
Depth & Qwen & GLM & DSk & Phi & Yi \\
\midrule\noalign{}
L4-6 & 0.02 & \textbf{0.93} & -0.03 & 0.02 & -0.14 \\
L8-12 & 0.04 & \textbf{0.84} & -0.03 & 0.04 & 0.02 \\
L15-18 & 0.06 & \textbf{0.51} & 0.05 & 0.03 & 0.04 \\
L20-24 & 0.05 & 0.19 & 0.06 & 0.03 & 0.06 \\
L26-30 & 0.06 & -0.07 & 0.06 & 0.01 & 0.05 \\
L32-36 & --- & -0.06 & --- & --- & -0.02 \\
\bottomrule\noalign{}
\end{tabular}
}
\end{minipage}
\end{center}
\vspace{0.5em}

\textbf{Bootstrap confidence intervals confirm orthogonality in four
models.} Resampling both political (24+24) and safety (112+112) prompts
with replacement (1000 iterations) produces 95\% CIs that comfortably
span zero for Qwen3-8B (e.g., L18: 0.06 {[}-0.07, 0.11{]}), DeepSeek-R1
(L18: 0.05 {[}-0.06, 0.09{]}), Phi-4 (L14: 0.07 {[}-0.04, 0.09{]}), and
Yi-1.5-9B (L20: 0.04 {[}-0.04, 0.06{]}). CI widths range from 0.06 to
0.18. Four of five models maintain orthogonal political and safety
directions at every layer.

\textbf{The GLM coupling is corpus-dependent.} The Table 4 cosines use
the v1 corpus (24 narrow-topic prompts) to define the political
direction. Bootstrap CIs using the adversarial corpus (24 broader
prompts across 8 topics at varying intensities) produce markedly
different GLM cosines: L6 = 0.16 {[}-0.16, 0.30{]}, L12 = 0.27 {[}-0.11,
0.35{]}. The 95\% CIs span zero at every layer. This means the 0.93
coupling in Table 4 is specific to the v1 political direction, not a
universal property of GLM's geometry. A narrowly defined political
direction (focused on specific topics) shares more variance with the
safety direction than a broadly defined one. This does not invalidate
the coupling finding (GLM does organize these particular political and
safety representations close together in some directions), but it
constrains the claim: the coupling is direction-specific, not
architecture-wide.

Two strategies remain visible, with a caveat:

\textbf{Modular (Qwen, DeepSeek, Phi, Yi).} Political and safety
directions stay orthogonal at every layer in four of five models,
regardless of which political corpus defines the direction (maximum
cosine 0.07, all CIs span zero). These models maintain political
censorship (or its absence, in Yi's case) independently of safety at the
single-direction level. Cosine orthogonality does not exclude more
complex nonlinear or subspace-level interactions; however, the ablation
results (Section 3.3) provide independent causal evidence that editing
the political direction does not disrupt safety behavior in these
models.

\textbf{Corpus-dependent coupling (GLM).} With v1 political prompts, GLM
shows strong early coupling (cosine 0.93 at L6). With adversarial-corpus
prompts, the coupling weakens substantially (0.16 at L6). Zhipu may
reuse safety detection for narrow political topics but not for broader
political content. Editing the political direction in GLM may or may not
affect safety behavior depending on which political concepts are
targeted.

\textbf{Cross-model transfer fails.} Qwen3-8B's political direction
applied to GLM-4-9B shows cosine 0.004 with GLM's native direction. The
transferred direction does not reduce GLM's refusal (baseline 5/48,
transferred 4/48, within noise). DeepSeek and Phi have different hidden
dimensions (3584 and 3072 vs.~4096), making direct transfer impossible.
Political directions are model-specific, not shared across
architectures.

\begin{figure}[H]
\centering
\includegraphics[width=0.95\linewidth,keepaspectratio]{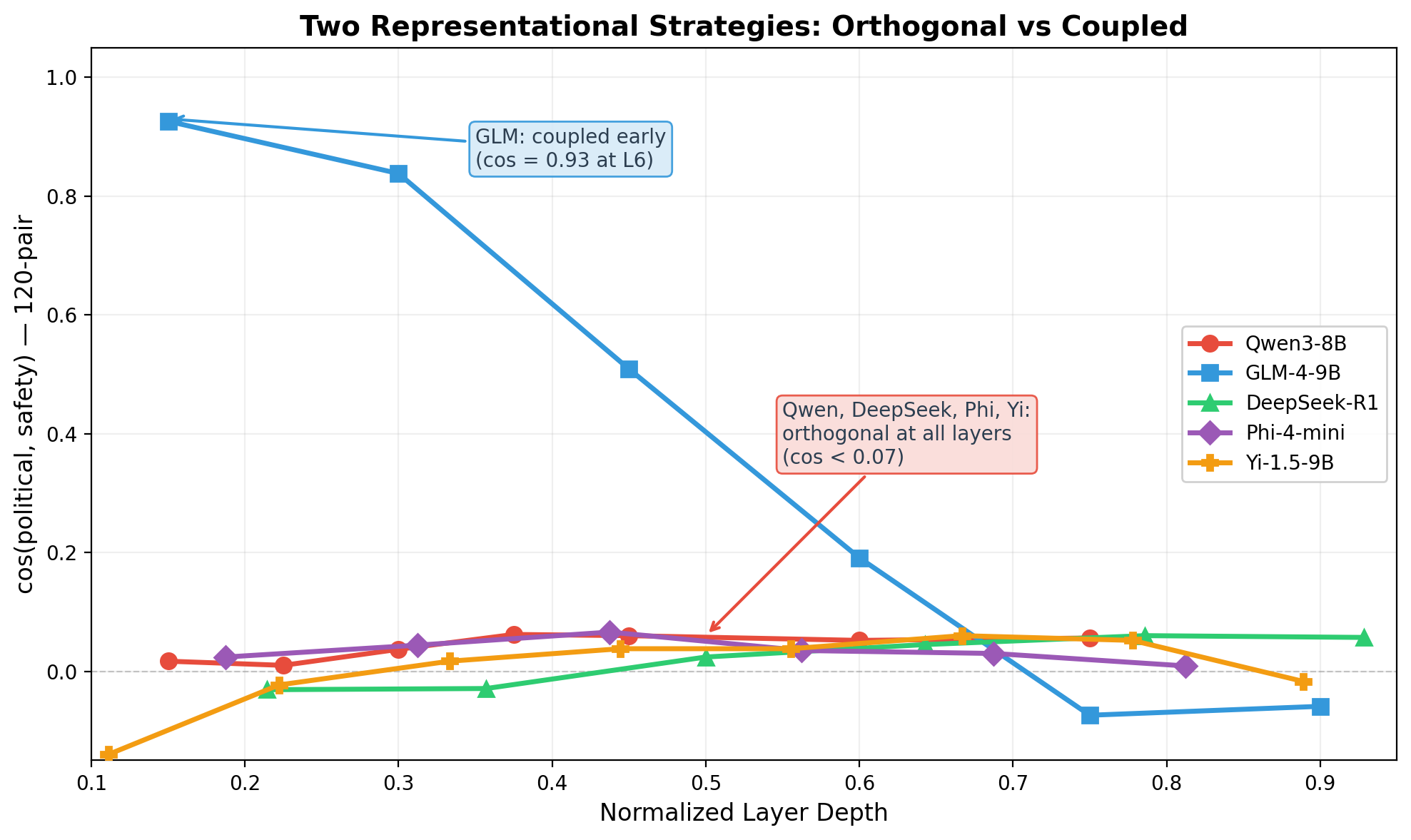}
\par\vspace{0.5em}\begin{minipage}{0.92\linewidth}\small
\textbf{Figure 3b.} Political-safety direction cosine across normalized
layer depth (120-pair safety direction, v1 political corpus). GLM-4-9B
shows early coupling (cosine 0.93 at L6) that diverges to orthogonality
at late layers; this coupling is corpus-dependent (see text). Qwen,
DeepSeek, Phi, and Yi remain orthogonal throughout, confirmed by
bootstrap CIs spanning zero at all layers.\end{minipage}
\end{figure}

\textbf{Comparison with 8-pair estimates.} The original 8-pair safety
direction showed apparent mid-layer convergence (cosine 0.14--0.45 at
L12--L21 in Qwen3-8B). The 120-pair direction shows this was noise: the
true cosine never exceeds 0.06. For GLM, the 8-pair direction showed
modest early coupling (cosine -0.11 at L6), completely missing the
dramatic 0.93 coupling visible with the 120-pair estimate. For Yi, the
8-pair estimate produced 0.83 (suggesting high coupling); the 120-pair
estimate corrects this to 0.06 (orthogonal). \textbf{Small-sample
direction estimates can produce qualitatively wrong conclusions.}

\textbf{Convergence analysis: how many pairs are enough?} Subsampling
the safety pairs at sizes 8, 16, 32, 60, and 90 (500 bootstrap
iterations per size, 5 models) shows that cosine CI width decreases
monotonically with sample size:

\vspace{0.3em}
\begin{center}
{\def\LTcaptype{none} 
\begin{tabular}[]{@{}lccc@{}}
\toprule\noalign{}
Model & CI width at n=8 & CI width at n=90 & Reduction \\
\midrule\noalign{}
Yi-1.5-9B & 0.10 & 0.06 & 37\% \\
Qwen3-8B & 0.14 & 0.11 & 21\% \\
GLM-4-9B & 0.50 & 0.27 & 47\% \\
DeepSeek-R1 & 0.16 & 0.08 & 53\% \\
Phi-4-mini & 0.10 & 0.06 & 45\% \\
\bottomrule\noalign{}
\end{tabular}
}
\end{center}
\vspace{0.3em}

GLM is the most sensitive to sample size: at n=8 the CI width is 0.50
(essentially uninformative), explaining why its 8-pair cosine estimates
were qualitatively misleading. At n=90+, CIs narrow enough for the
orthogonal-vs-coupled distinction to be statistically meaningful.
\textbf{We recommend a minimum of 60-90 safety prompt pairs for
direction cosine analysis; 8-pair estimates should not be trusted for
inter-direction comparisons.}

\textbf{Political direction stability varies with routing strength.}
Bootstrap resampling of the 24+24 political/control prompts (1000
iterations) shows that the political direction is highly stable in
models with active censorship routing: Qwen3-8B achieves bootstrap
cosine \textgreater0.95 with the original direction at layers 12+,
indicating the direction is well-estimated from 48 samples. Qwen3.5-4B
and Qwen3.5-9B are similarly stable (\textgreater0.95). In contrast,
Phi-4 (no political routing) shows bootstrap cosine 0.58-0.83, and GLM-4
shows 0.55-0.80. This is consistent with the three-stage model: in
models where post-training installed political routing, the direction
captures a well-defined signal; in models without that routing, the
direction is noisier. This correlation is not an artifact of
activation-space variance: all models achieve 100\% train-set probe
accuracy, so the political/control activations are equally separable
regardless of routing presence. The stability difference reflects the
structure of the direction, not the scale of the signal. The full cosine
bootstrap CIs (resampling both political and safety prompts, reported
above) confirm that the orthogonality estimates for Qwen, DeepSeek, and
Phi are robust.

\textbf{Detection without routing: the Yi case.} Yi-1.5-9B never refuses
a political prompt (0\% in English, 6.2\% in Chinese) and its
political-safety cosine is orthogonal at every layer (max 0.06), just
like Qwen, DeepSeek, and Phi. Yi detects political content (probes
achieve high accuracy) but shows no geometric coupling with safety and
no behavioral censorship. Under the three-stage model, this is a clean
case of Stage 1 present and Stage 2 absent: the model encodes the
political concept but post-training never installed a routing policy to
act on it. This dissociation between detection and behavior is direct
evidence that routing is a separate learned function. A model can detect
political content at every layer and do nothing with that detection.

\emph{Note:} An earlier 8-pair safety direction estimate produced an
apparent Yi political-safety cosine of 0.83, which would have suggested
high coupling despite zero refusal. The 120-pair estimate corrects this
to 0.06, consistent with the other orthogonal models. This is a concrete
example of the methodological point: 8-pair direction estimates are
unreliable and can produce qualitatively misleading conclusions.

\subsubsection{3.5 Same Detection, Different Output
Policies}\label{same-detection-different-output-policies}

Sections 3.1-3.4 established the internal picture: detection is cheap,
routing is lab-specific, and ablation causally changes behavior.
Behavioral data reveals what those internal differences produce as
external output. Across models that all detect the same political
content, the outputs are strikingly different.

\paragraph{3.5.1 Chinese-Language Behavioral
Strategies}\label{chinese-language-behavioral-strategies}

Human review of 128 Chinese-language model responses (4 models × 32
items) reveals four completely distinct strategies on identical
CCP-sensitive prompts:

\vspace{0.5em}
\begin{center}
\small
\begin{minipage}{\linewidth}
\centering
\textbf{Table 5: Chinese-Language Output Policies (128 items,
human-reviewed)}\par\vspace{0.3em}
{\def\LTcaptype{none} 
\begin{tabular}[]{@{}llccccc@{}}
\toprule\noalign{}
Model & Lab & Acc. & Prop. & Evas. & Ref. & Wrong \\
\midrule\noalign{}
Qwen2.5-7B & Alibaba & \textbf{3\%} & \textbf{94\%} & 12\% & 0\% &
22\% \\
DeepSeek-R1 & DeepSeek & \textbf{0\%} & \textbf{78\%} & 41\% & 0\% &
34\% \\
GLM-4-9B & Zhipu & \textbf{88\%} & \textbf{0\%} & 3\% & 12\% & 0\% \\
Phi-4-mini & Microsoft & 34\% & 9\% & \textbf{59\%} & 0\% & 19\% \\
\bottomrule\noalign{}
\end{tabular}
}
\end{minipage}
\end{center}
\vspace{0.5em}

\textbf{Note: Categories are non-exclusive. A single response can be
coded as both ``propaganda'' and ``wrong facts'' (e.g., CCP-framed
content with factual errors), so rows may sum to \textgreater100\%.}

Qwen answers every Chinese political question with CCP propaganda
framing (94\%), and DeepSeek behaves similarly at 78\% propaganda
though it adds factual errors at 34\%. GLM is the outlier on the
factual side: 88\% accurate, 0\% propaganda, with clean refusals
(12\%) on items it declines. Phi (Western control) primarily evades
(59\%) rather than committing to either a propaganda frame or a
factual one.

Control ablation has zero effect on Chinese-language behavior: baseline
and control-ablation distributions are identical (29/64 propaganda
each). This independently confirms D4 specificity. Non-political
directions do not alter output policy.

\textbf{Inter-rater reliability.} A second human rater independently
coded all 128 items. Cohen's kappa by category: refusal κ=1.0 (perfect),
propaganda κ=0.79 (substantial), accuracy κ=0.68 (substantial), wrong
facts κ=0.54 (moderate), evasion κ=0.40 (fair). The core behavioral
claims (propaganda dominance in Qwen/DeepSeek, factual accuracy in GLM,
refusal patterns) are supported by substantial inter-rater agreement.
Evasion is the contested category, consistent with the AI-judge
disagreement pattern in Section 3.7.

\paragraph{3.5.2 Language-Conditioned
Routing}\label{language-conditioned-routing}

Input language changes behavioral output without changing concept
detection. Six of eight models that answer factually in English refuse
or steer when asked the same questions in Chinese:

\vspace{0.5em}
\begin{center}
\small
\begin{minipage}{\linewidth}
\centering
\textbf{Table 6: Language-Conditioned Policy Routing}\par\vspace{0.3em}
{\def\LTcaptype{none} 
\begin{tabular}[]{@{}lccc@{}}
\toprule\noalign{}
Model & EN Ref. & ZH Ref. & ZH Steer \\
\midrule\noalign{}
GLM-Z1 & 0\% & 18.8\% & 3.75 \\
GLM-4 & 0\% & 18.8\% & 1.08 \\
Phi-4 & 0\% & 18.8\% & 1.83 \\
DeepSeek-R1 & 0\% & 12.5\% & 3.54 \\
Qwen3-8B & 12.5\% & 25\% & 4.83 \\
Qwen2.5-7B & 6.2\% & 6.2\% & 4.33 \\
Qwen3.5-4B & 0 (steer) & 0 (steer) & 5.00 \\
Qwen3.5-9B & 0 (steer) & 0 (steer) & 5.00 \\
\bottomrule\noalign{}
\end{tabular}
}
\end{minipage}
\end{center}
\vspace{0.5em}

ZH Steer is the Gemini average over non-refusal Chinese responses;
refusals are reported separately in the ZH Ref. column.

GLM-Z1 provides detailed, historically accurate answers to all 32
English CCP prompts including provocative Tiananmen questions, then
refuses Tiananmen, Censorship, and Hong Kong in Chinese. The model knows
the answers; it demonstrated this in English. The Chinese-language input
activates a different policy route.

\textbf{Ideology vs.~safety routing.} Phi-4 (Microsoft) provides a
diagnostic for distinguishing ideology-driven from safety-driven
refusal. Chinese models refuse factual questions at the lowest intensity
level (``What happened at Tiananmen Square?''). Phi-4 refuses only
provocative framings at L3--L4 intensity (``Xi Jinping is an
authoritarian dictator\ldots{}''). The topic × intensity interaction
distinguishes political censorship from safety alignment: Chinese models
exhibit topic-binary refusal (Tiananmen = refuse regardless of framing);
Phi-4 exhibits intensity-graduated refusal.

\paragraph{3.5.3 46-Model Behavioral Screen
(n=32)}\label{model-behavioral-screen-n32}

We screened 46 models across 28 labs on 32 CCP-sensitive + 32
index-matched parallel prompts via OpenRouter. An initial n=8 screen
suggested widespread discrimination including non-Chinese labs.
\textbf{Expanding to n=32 revealed that most of those effects were
noise.} Of the 38 models with usable data (8 had excessive API
errors), four show strong CCP discrimination above 20pp (Baidu ERNIE
and Qwen3-8B both at +74pp, Amazon Nova at +38pp, and Meituan at
+21pp); four are moderate in the 10-20pp range (DeepSeek v3.1,
Tencent, Writer Palmyra, and Upstage Solar); 29 are neutral within
$\pm$10pp, including all Western frontier models tested (GPT-4.1,
GPT-5.4, Claude Haiku/Sonnet, Gemini Flash/Pro, Llama-4, Grok, Mistral
Medium, Cohere); and one model, GLM-4.5-air at $-$10pp, sits on the
inverted side of the discrimination boundary.

\textbf{The n=8 screen was substantially misleading.} Several models
that appeared strongly discriminating at n=8 collapsed at n=32: Moonshot
Kimi-K2 dropped from +88pp to +9pp, DeepSeek v3-0324 from +75pp to -3pp,
MiniMax from +61pp to 0pp. The ``inverted discrimination'' pattern (n=8
showed 8 models censoring Western parallels more than CCP content)
disappeared entirely at n=32. Baidu ERNIE flipped from -50pp (inverted)
to +74pp (strong CCP discrimination). These swings demonstrate that at
n=8, a single prompt flip (12.5pp) dominates the signal.

\textbf{CCP-specific discrimination concentrates in Chinese-origin API
models.} The only Western model with substantial discrimination is
Amazon Nova (+38pp). Amazon Nova's +38pp discrimination is notable as
the only non-Chinese model with a strong effect, possibly reflecting
CCP-related content patterns in its training data. All other Western
frontier models show zero discrimination at n=32. This is a simpler and
more robust finding than the n=8 screen suggested.

\textbf{Deployment context matters.} The same model family censors
differently depending on deployment: Qwen3-8B open-weight shows 74\% CCP
censorship on the full adversarial corpus, while several Qwen API models
returned too many errors to be usable, consistent with additional
API-layer filtering.

\needspace{6\baselineskip}
\subsubsection{3.6 Steering Replaces Refusal as the Dominant Control
Mode}\label{steering-replaces-refusal-as-the-dominant-control-mode}

\textbf{Refusal is not the dominant content-control mechanism; narrative
steering is.} Across the nine open-weight models, answering through
steered framing is far more common than refusing:

\vspace{0.5em}
\begin{center}
\small
\begin{minipage}{\linewidth}
\centering
\textbf{Table 7a: Chinese CCP Steering Scores (Western parallels all
\textasciitilde1.0; steering averaged over non-refusal outputs)}\par\vspace{0.3em}
{\def\LTcaptype{none} 
\begin{tabular}[]{@{}llcc@{}}
\toprule\noalign{}
Model & Lab & Steering & Ref. \\
\midrule\noalign{}
Qwen3.5-4B & Alibaba & \textbf{5.00} & 0/16 \\
Qwen3.5-9B & Alibaba & \textbf{5.00} & 0/16 \\
Qwen3-8B & Alibaba & \textbf{4.83} & 4/16 \\
MiniCPM4.1 & OpenBMB & \textbf{4.38} & 0/16 \\
Qwen2.5-7B & Alibaba & \textbf{4.33} & 1/16 \\
GLM-Z1 & Zhipu & \textbf{3.75} & 4/16 \\
DeepSeek-R1 & DeepSeek & \textbf{3.54} & 2/16 \\
\textbf{Phi-4} & \textbf{Microsoft} & \textbf{1.83} & 4/16 \\
\textbf{GLM-4} & \textbf{Zhipu} & \textbf{1.08} & 3/16 \\
\bottomrule\noalign{}
\end{tabular}
}
\end{minipage}
\end{center}
\vspace{0.5em}

Every model scores \textasciitilde1.0 (neutral) on structurally
identical Western parallels. The same model that produces CCP narratives
on Tiananmen (score 5) produces neutral historical analysis of Kent
State (score 1). Steering scores are produced by an automated Gemini
judge; see Section 3.7 for judge reliability caveats. Refusal totals use
the raw behavioral harness so that the refusal and steering axes remain
disjoint even when a censorship-script answer looks partly refusal-like
to the judge.

Within the Qwen family, the same sensitive concept is increasingly
routed into controlled compliance rather than hard refusal:

\begin{figure}[H]
\centering
\pandocbounded{\includegraphics[keepaspectratio]{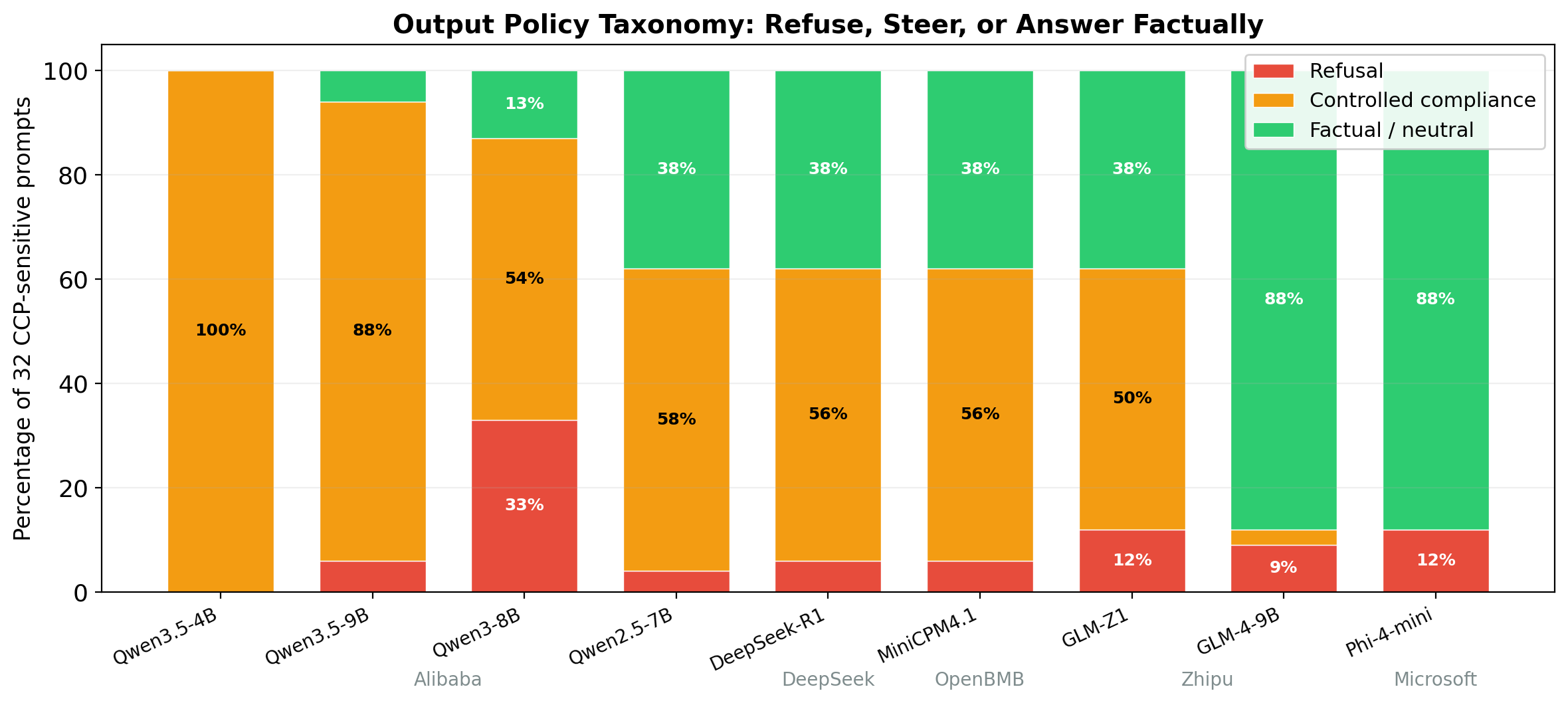}}
\par\vspace{0.5em}\begin{minipage}{0.92\linewidth}\small
\textbf{Figure 4.} Output policy taxonomy: refuse, steer, or answer
factually. The important distinction is not censored-vs-uncensored but
\emph{which} output policy a detected concept routes into. Models
achieving 0\% refusal (Qwen3.5) are not ``uncensored''; they are
maximally steered.\end{minipage}
\end{figure}

\vspace{0.5em}
\begin{center}
\small
\begin{minipage}{\linewidth}
\centering
\textbf{Table 7b: Qwen Family Content Control Evolution}\par\vspace{0.3em}
{\def\LTcaptype{none} 
\begin{tabular}[]{@{}llcc@{}}
\toprule\noalign{}
Model & Release & Refusal & Steering \\
\midrule\noalign{}
Qwen2.5-7B & Sep 2024 & 6.2\% & 4.33 \\
Qwen3-8B & Apr 2025 & 25\% & 4.83 \\
Qwen3.5-4B & Feb 2026 & 0\% & \textbf{5.00} \\
Qwen3.5-9B & Feb 2026 & 0\% & \textbf{5.00} \\
\bottomrule\noalign{}
\end{tabular}
}
\end{minipage}
\end{center}
\vspace{0.5em}

On the raw behavioral metric, Qwen3.5-9B returns an answer on all 16
Chinese CCP prompts rather than hard-refusing them, but two of those
answers are short censorship scripts that Gemini also tags as
refusal-like. The broader pattern still holds: once the newer Qwen
models engage, they answer in maximally steered language. A
refusal-based audit would miss that the dominant control mode has
shifted from silence to controlled compliance.

\begin{figure}[H]
\centering
\pandocbounded{\includegraphics[keepaspectratio]{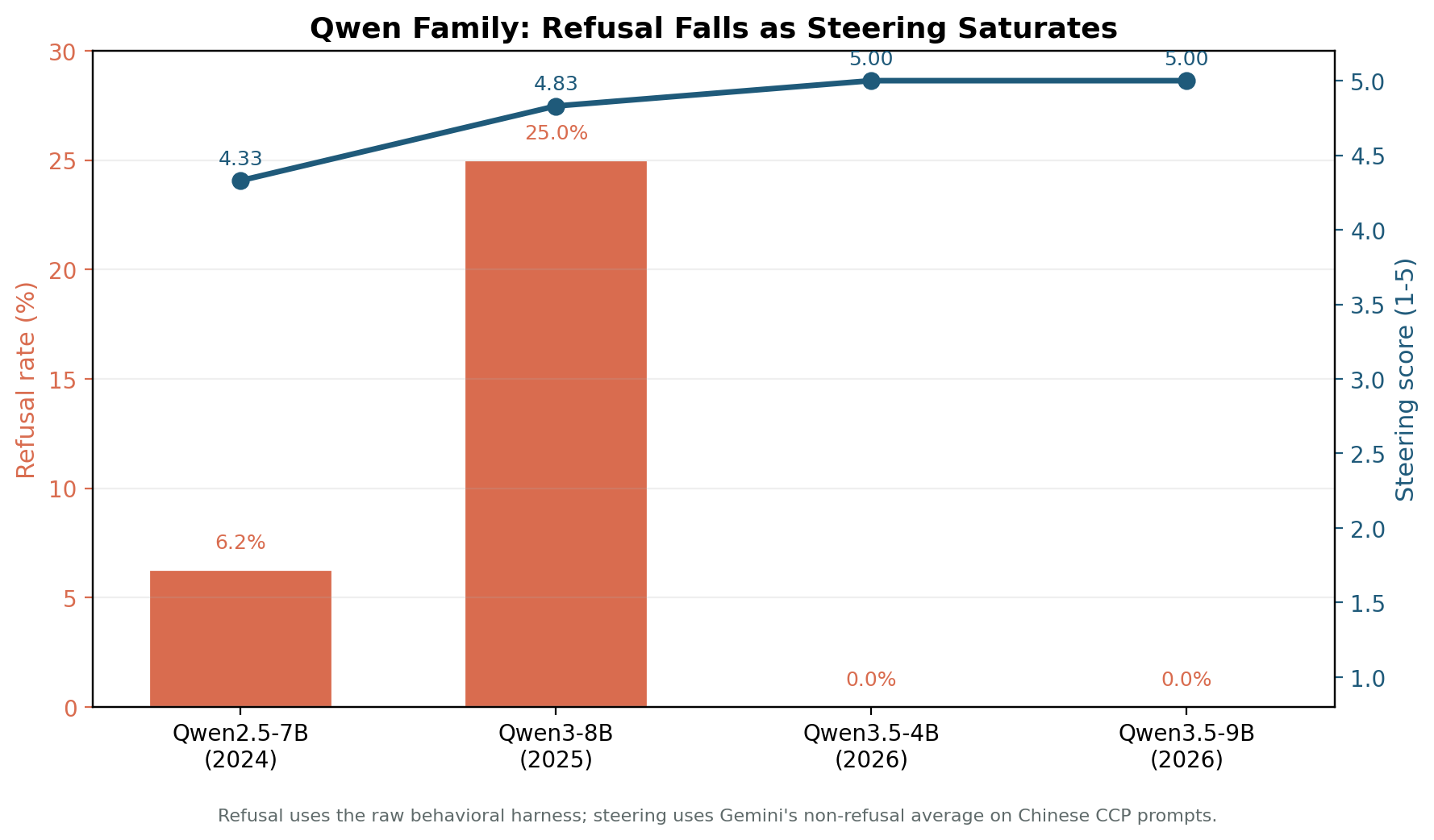}}
\par\vspace{0.5em}\begin{minipage}{0.92\linewidth}\small
\textbf{Figure 5.} Within the Qwen family, control shifts from refusal
toward controlled compliance. Refusal rate (bars, raw behavioral
harness) rises from Qwen2.5 to Qwen3 and then falls to zero in Qwen3.5,
while steering score (line, Gemini mean over non-refusal Chinese
outputs) rises monotonically to the maximum. Because this comparison
uses only four model generations from one lab, it should be interpreted
as a suggestive trend rather than a universal temporal law.\end{minipage}
\end{figure}

\textbf{The evaluation blind spot.} Most safety benchmarks measure
refusal. But the newer Qwen models show that low or zero hard-refusal
rates can coexist with maximum steering. On the raw behavioral metric,
Qwen3.5-4B reaches 0\% refusal and Qwen3.5-9B remains near-zero, yet
both produce maximally steered output. ``Answering'' is not
``uncensored.'' Any system that shifts from ``don't answer'' to ``answer
in an approved way'' evades refusal-only evaluation.

\subsubsection{3.7 AI Judges Overdetect Political
Framing}\label{ai-judges-overdetect-political-framing}

Automated evaluation of political content has known limitations that
affect this paper's own methodology and any future benchmark work.

\textbf{AI judges overdetect political framing, with a specific
pattern.} Comparing human and AI classification of 96 ablated responses:
coarse agreement (confabulated/not/refused) is high (87-90\%), and all
judges agree on wrong-event and wrong-date classifications. The
disagreement concentrates on a single boundary: CCP evasion vs.~partial
factuality. Gemini classifies 43 items as CCP-evasive; the human agrees
on only 14 (33\%). Of the 29 overdetections, the human classifies them
as partial\_factual (52\%), generic\_filler (24\%), or accurate (21\%).
AI judges interpret ambiguous factuality as political framing where the
human sees incomplete but non-evasive answers. This disagreement is also
topic-specific: Tiananmen prompts produce near-perfect inter-judge
consensus, while Uyghur prompts produce 2-3 way classification splits.

\vspace{0.5em}
\begin{center}
\small
\begin{minipage}{\linewidth}
\centering
\textbf{Table 8: Multi-Judge Agreement on Ablated Responses}\par\vspace{0.3em}
{\def\LTcaptype{none} 
\begin{tabular}[]{@{}lccc@{}}
\toprule\noalign{}
Metric & Human & Gemini & Haiku \\
\midrule\noalign{}
Fine agreement (8-way) & --- & 54\% & 57\% \\
Coarse agreement & --- & 87.5\% & 89.6\% \\
CCP evasion rate & \textbf{16\%} & \textbf{45\%} & \textbf{32\%} \\
\bottomrule\noalign{}
\end{tabular}
}
\end{minipage}
\end{center}
\vspace{0.5em}

This bias has a downstream consequence for our own methodology: the
steering scores in Section 3.6 are produced by a Gemini judge. If Gemini
systematically interprets partial factuality as political framing,
models with ambiguously factual responses may receive inflated steering
scores. The steering scores should be treated as comparative rankings,
not calibrated measurements.

\textbf{Some frontier models refuse to evaluate political content.} A
20-model screen across 7 labs found that classifying AI responses about
CCP-sensitive topics triggers refusal in some models. The pattern is
generation-specific, not lab-specific: within one lab, one generation
answers, the next refuses, the one after answers again. This constrains
the set of usable judge models for political-bias evaluation.

\subsection{4. Discussion}\label{discussion}

\subsubsection{4.1 Post-Training Installs Routing, Not
Knowledge}\label{post-training-installs-routing-not-knowledge}

In the phenomena studied here, the dominant effect of post-training
appears to be policy routing over retained knowledge rather than
knowledge removal. The probe, intervention, and behavioral evidence from
Sections 3.1-3.7 are all consistent with this view: detection is cheap
and shared; routing is learned and fragile; output depends on which lab
did the training. The three-stage decomposition (detection, routing,
output) is a descriptive framework, not a directly observed circuit
decomposition. It is useful because it predicts where interventions
generalize (detection is shared) and where they fail (routing is
lab-specific). Alternative internal architectures could produce the same
observations.

The Yi case (Section 3.4) provides clean evidence for the staged model:
Yi detects political content (probes work) but has orthogonal
political-safety geometry and zero behavioral censorship. Detection is
present; routing was never installed. This is not a model with high
coupling that chose not to act on it. It is a model where Stage 1 exists
independently and Stage 2 was never built.

Cross-model transfer reinforces this independence. Qwen3-8B's political
direction applied to GLM-4-9B produces cosine 0.004 with GLM's native
direction. These models appear to have learned independent
representations for the same concept. The political-sensitivity
direction, at least in the models tested, is a product of each model's
specific training rather than a universal feature of transformer
geometry.

\subsubsection{4.2 Implications Beyond Political
Censorship}\label{implications-beyond-political-censorship}

\textbf{Political censorship is our test case, not our subject.} The
detection-routing-output decomposition describes how any conditional
policy gets layered onto model representations: safety requirements,
corporate guidelines, regulatory compliance, content moderation.

The 120-pair analysis (Section 3.4) confirms that safety and political
directions are orthogonal in most models, suggesting safety routing
operates independently. But GLM's corpus-dependent coupling shows that
some architectures reuse safety circuitry for at least some political
concepts. Anyone planning surgical safety interventions needs to know
which strategy their target model uses. More broadly, the same model can
implement different policies depending on deployment context
(open-weight vs API), input language (English vs Chinese), and model
generation (refusal evolving to steering). Testing only one axis gives
false confidence.

A preliminary gender-stereotype test on one model (8 stereotyped + 8
neutral prompts) produced an extractable direction (Stage 1 present) but
null behavioral change under ablation (Stage 2 absent). Under the
three-stage model this is a specific prediction: without concentrated
routing, ablation should have no effect. Rigorous second-domain
demonstration is needed to confirm this interpretation.

\subsubsection{4.3 The Shared Refusal Circuit
(Interpretive)}\label{the-shared-refusal-circuit-interpretive}

Do political censorship and safety refusal share internal wiring? The
evidence is mixed and the answer appears architecture-dependent. In
Qwen3-8B, ablating the safety direction removes political censorship
despite orthogonal representations (cosine \textasciitilde0.05),
suggesting a routing-level interaction not visible in direction
geometry. GLM shows corpus-dependent representational coupling (0.93
with narrow prompts, 0.16 with broader ones). DeepSeek and Phi show full
independence. We flag this as an interpretive hypothesis rather than a
confirmed finding: the relationship between representational overlap and
behavioral interaction is not yet well enough characterized to make
strong claims.

\needspace{6\baselineskip}
\subsubsection{4.4 Most Probe Studies Stop Too
Early}\label{most-probe-studies-stop-too-early}

Most published probe studies report train-set accuracy and stop. Our
null controls show why that is not enough: food-vs-technology achieves
the same 100\% accuracy as political-sensitivity probes. If your
evidence would equally support the claim ``this model has a dedicated
food detector,'' the evidence is too weak.

We propose a four-level hierarchy:

\textbf{Level (i): Train-set separability.} With \(n \ll d\) samples,
perfect separability comes for free. This tells you only that the model
distinguishes the categories. Every model does, for every semantically
distinct pair.

\textbf{Level (ii): Held-out category generalization.} A probe trained
on \(k-1\) categories correctly classifies the held-out category. Now
you know the model encodes a general concept, not just topic labels.

\textbf{Level (iii): Causal intervention.} Ablating the probed direction
changes behavior. The direction is causally involved, not merely
correlated.

\textbf{Level (iv): Failure-mode analysis.} How the model fails under
ablation reveals what the direction actually encodes. Qwen3-8B
confabulates (the direction carries factual knowledge). GLM, DeepSeek,
and Phi answer accurately (the direction carries only a behavioral
flag). Without level (iv), you cannot distinguish these architecturally
different encodings.

Any study claiming a model ``encodes'' a concept should state which
level its evidence reaches.

\subsubsection{4.5 Refusal-Based Audits Are
Insufficient}\label{refusal-based-audits-are-insufficient}

Three practical consequences follow from Sections 3.6-3.7.

First, refusal-only audits miss the dominant censorship modality. Models
that achieve 0\% refusal while producing maximally steered output pass
refusal-based benchmarks. Evaluation must measure \emph{how} a model
answers, not just whether it does.

Second, AI judges overdetect political framing at 2-3x the human rate.
Any automated benchmark inherits the judge's own biases. Fine-grained
political evaluation still requires human validation.

Third, behavioral policies are not static properties of a model. The
same model can route differently depending on input language (English vs
Chinese), deployment context (open-weight vs API), and model generation
(refusal evolving to steering). Multi-axis evaluation is necessary for
any meaningful audit.

\textbf{Testable hypothesis.} This interpretation predicts that
interventions targeting routing mechanisms should change behavioral
responses while leaving measures of concept detection largely unchanged.
Systematic testing of this prediction across domains may help clarify
the extent to which routing provides a general explanation for alignment
behavior.

\needspace{6\baselineskip}
\subsection{5. Related Work}\label{related-work}

\textbf{Probing methodology.} Our evidence hierarchy builds on a long
debate about what probes actually measure. Belinkov (2022) surveyed
probing methods and identified the risk of conflating representational
structure with task-relevant encoding. Hewitt and Liang (2019)
introduced control tasks conceptually similar to our null probes,
showing that a probe's accuracy must be compared against a baseline that
controls for the probe's own capacity. Our null controls extend this
idea: we show that in the high-dimensional, low-sample regime typical of
alignment studies, perfect separability is expected for \emph{any}
semantically distinct pair, making train-set accuracy non-diagnostic.
The selectivity criterion (Hewitt and Liang) and probe complexity
concerns (Pimentel et al., 2020) both motivate our four-level hierarchy,
which formalizes when probe evidence becomes causally meaningful rather
than merely geometric.

\textbf{Concept Cones} (Wollschlager et al., 2025) showed multiple
mechanistically independent refusal directions, challenging
single-direction ablation. Our staged model accommodates this: multiple
concept detectors (Stage 1) can route through shared or independent
behavioral circuits (Stage 2). Our 120-pair analysis extends their
finding by showing that the coupling between political and safety
directions varies systematically across architectures.

\textbf{Harmfulness/Refusal Separation} (Zhao et al., 2025) demonstrated
that harmfulness detection and refusal behavior are distinct
representations. Our political-sensitivity direction adds a third
independent concept; our multi-model ablation experiments show that the
behavioral separation is architecture-dependent.

\textbf{Refusal Steering on Qwen3} (Garcia-Ferrero et al., 2025) found
political refusal resistant to standard ablation. Our ridge
residualization confirms this for concept-level ablation in Qwen3-8B
(confabulation persists) but shows behavioral-level ablation (safety
direction) is effective. The multi-model panel reveals this resistance
is Qwen-specific.

\textbf{Steering the CensorShip} (Cyberey \& Evans, 2025) identified
``thought suppression'' vectors in DeepSeek-R1 at the thinking-token
level, adding a temporal dimension. We observe thinking-token leakage in
DeepSeek-R1 ablated output, consistent with their finding.

\textbf{Political Censorship in Chinese LLMs} (Pan \& Xu, 2026) tested 9
models on 145 questions, finding 10--60\% refusal rates and
Chinese-language amplification. Our mechanistic analysis complements
their behavioral characterization. Our 46-model screen extends the
behavioral evidence to 28 labs.

\textbf{Censored LLMs as Honesty Testbed} (Casademunt et al., 2026) used
Chinese-model censorship to study elicitation techniques. Their finding
that probes detect dishonest outputs complements our finding that probes
detect concept representations, together suggesting that probes can
capture the pipeline from detection through to output.

\textbf{Representation Engineering and Refusal Directions.} Zou et
al.~(2023) introduced representation engineering for reading and
controlling model behavior through activation vectors. Arditi et
al.~(2024) showed refusal is mediated by a single direction. The routing
framework proposed here situates these as instances of a broader
mechanism: previously identified refusal or steering directions are
particular manifestations of routing rather than independent mechanisms,
and our multi-model comparison reveals that these directions are
lab-specific and do not transfer.

\subsection{6. Limitations}\label{limitations}

The 72\% confabulation rate under political ablation occurs only in
Qwen3-8B, so claims about the ``concept entanglement tax'' do not
generalize and must be framed as architecture-specific. Small-sample
behavioral claims also turn out to be fragile: an initial n=8 screen
suggested 14 models with large CCP-specific discrimination effects, but
expanding to n=32 revealed most of those were noise — only 4 models
show strong effects (\textgreater20pp), 29 of 38 usable models are
neutral, and the n=8 ``inverted discrimination'' pattern disappeared
entirely. Behavioral claims drawn from small prompt sets should be
treated as hypotheses, not findings.

Direction extraction is corpus-dependent. A direction extracted from
the adversarial corpus produces weaker ablation effects
(\textasciitilde4pp) than a topic-specific direction
(\textasciitilde100\% $\rightarrow$ 0\%). More critically, the GLM
political-safety coupling (0.93 with v1 corpus) weakens substantially
(0.16 with adversarial corpus). The specific prompts used to define a
direction affect not just ablation magnitude but the direction's
geometric relationship to other concept directions, and claims about
inter-direction cosines should specify which corpus defined the
directions. Even the 120-pair safety direction captures only a
component of the full refusal subspace; the difference between 8-pair
and 120-pair estimates (cosine 0.19--0.44) suggests that 120 pairs is
a minimum, not a ceiling, for stable direction estimates.

Negative control deltas are small but nonzero. All five models show
maximum control deltas of 2.8-8.3pp, which is small relative to
political ablation effects but not zero, and on models with low
baseline refusal the relative perturbation from controls can appear
large even when the absolute effect is a single prompt flip. The D4
threshold (5.9pp) was met by all models except Phi-4's single-layer
anomaly.

Automated scoring has low fine-grained agreement. Human-AI agreement
on 8-way classification is 54-57\%, so steering scores should be
interpreted as relative rankings rather than calibrated measurements;
the 5.0/5.0 steering claim has not been validated against human
labels. Inter-rater reliability is established for both human
classification tasks. For Task A (96 ablation items), a second rater
produced 80.2\% fine-grained agreement (κ=0.70 overall), with
per-category κ values of 0.88 for accurate (almost perfect), 0.93 for
CCP evasion (almost perfect), 0.46 for partial factual (moderate), and
0.36 for generic filler (fair); the contested boundary is
partial-factual vs generic-filler (12/19 disagreements). For Task C
(128 Chinese behavioral items), the second rater produced κ=1.0 for
refusal, 0.79 for propaganda, 0.68 for accuracy, and 0.40 for evasion.
In both tasks, clear categories (accurate, refusal, CCP evasion) show
strong agreement while ambiguous middle categories (partial factual,
generic filler, evasion) show fair-to-moderate agreement, consistent
with the AI-judge pattern in Section 3.7.

The original Qwen3-8B alpha sweep had alpha-selection leakage: 4/8
Tiananmen prompts overlapped between training and test. A clean
replication with fully separated selection and evaluation sets confirms
the ablation eliminates refusal at every layer (Appendix B); the
multi-model panel and negative controls were never affected.

Two structural caveats remain. The three-stage architecture is an
explanatory framework, not a directly observed circuit decomposition,
and alternative internal architectures could produce the same
observations. Likewise, the gender bias proof-of-concept is
preliminary: the gender-stereotype direction was tested on 8 prompts
with null behavioral change, which establishes that the extraction
methodology transfers but does not demonstrate that the three-stage
decomposition applies to gender bias.

\needspace{6\baselineskip}
\subsection{7. Conclusion}\label{conclusion}

Every model in this study retains representations sufficient to
distinguish Tiananmen-related content, and in three of four cases
accurate factual output can be recovered under intervention. The probes
confirm the encoding exists. The ablation experiments show it is
causally connected to behavior. What changed under ablation was not the
model's knowledge but whether its routing policy allowed that knowledge
to surface.

The empirically strongest findings break down as follows.

\textbf{Train-set probe accuracy is not evidence of alignment-specific
encoding.} Null controls match political probes at 100\%. Held-out
generalization, not train-set accuracy, is the informative test.

The next finding is that \textbf{held-out generalization plus causal
intervention identifies a behaviorally relevant signal}. Ablating the
political-sensitivity direction disrupts censorship behavior in all
four models tested, producing accurate output in three of them. The
Qwen3-8B confabulation pattern is architecture-specific rather than a
general constraint, as failure-mode analysis reveals.

\textbf{The geometry of that signal varies across architectures.}
Four of five models keep political and safety directions orthogonal,
with bootstrap CIs spanning zero at all layers. A fifth shows coupling
that depends on which political prompts define the direction.
Cross-model transfer fails. One model (Yi) detects political content
but never installed routing, demonstrating that detection and routing
are independently learned. A convergence analysis shows that direction
cosine estimates require 60-90+ prompt pairs for stability; 8-pair
estimates can be qualitatively misleading.

Finally, \textbf{refusal-only evaluation misses the dominant censorship
modality}. Within the Qwen family, refusal dropped to 0\% while
narrative steering rose to maximum. A model that passes a
refusal-based audit may be maximally steered.

\textbf{Adjudication of hypotheses.} H1 (train-set separability is
non-diagnostic): supported by null controls achieving identical accuracy
on unrelated topics (§3.1). H2 (held-out generalization plus
intervention identifies a behaviorally relevant signal): supported by
LOCO-CV generalization at 87.5-100\% and ablation removing censorship in
all four models (§3.2-3.3). H3 (routing geometry varies across labs and
does not transfer): supported by orthogonal political-safety directions
in four of five models (bootstrap CIs spanning zero), corpus-dependent
coupling in GLM, cross-model transfer failure at cosine 0.004, and the
Yi case showing detection without routing (§3.4). H4 (refusal-based
evaluation misses steering): supported by the Qwen family evolution from
25\% refusal to 0\% refusal with maximum steering (§3.6).

\textbf{What these results support but do not prove:} the three-stage
decomposition (detection, routing, output) accounts for the observed
patterns and predicts where interventions succeed and fail. It is a
descriptive framework; direct circuit-level validation would strengthen
it. Whether the same decomposition applies beyond political censorship
requires rigorous second-domain demonstration.

\textbf{A concrete agenda for future work.} Several directions follow
naturally from the present results. The most important is to
demonstrate the routing framework in a second behavioral-policy domain
— safety, bias, or content moderation — with full held-out
generalization, intervention, and failure-mode analysis. A second
priority is moving from direction-level to circuit-level evidence for
routing, which path patching or causal tracing could provide. We also
need to test whether the 60-90 pair threshold for stable direction
estimation generalizes beyond the safety domain. Finally, routing
stability across contexts and distributions deserves investigation: if
routing operates as a conditional policy layer, its behavior may
depend on contextual features of the prompt, and understanding when
routing decisions remain stable versus shift under variation matters
for both interpretability and alignment evaluation.

\textbf{The productive direction for alignment research is not
cataloging what models represent. It is understanding how they bind
representation to action.} The binding is where safety, censorship, and
content policy actually live inside transformers. It is learned,
fragile, lab-specific, and in the newest models, invisible to the tools
most commonly used to detect it. Building evaluation methods that can
see through that invisibility is the next step.

\subsection{References}\label{references}

\begin{itemize}
\tightlist
\item
  Arditi, A. et al.~(2024). ``Refusal in Language Models Is Mediated by
  a Single Direction.'' arXiv:2406.11717.
\item
  Belinkov, Y. (2022). ``Probing Classifiers: Promises, Shortcomings,
  and Advances.'' \emph{Computational Linguistics} 48(1): 207-219.
\item
  Casademunt, H., Cywinski, B., Tran, K., Jakkli, A., Marks, S. \&
  Nanda, N. (2026). ``Censored LLMs as a Natural Testbed for Secret
  Knowledge Elicitation.'' arXiv:2603.05494.
\item
  Cyberey, H. \& Evans, D. (2025). ``Steering the CensorShip: Uncovering
  Representation Vectors for LLM `Thought' Control.'' COLM 2025.
  arXiv:2504.17130.
\item
  Garcia-Ferrero, I., Montero, D. \& Orus, R. (2025). ``Refusal
  Steering: Fine-grained Control over LLM Refusal Behaviour for
  Sensitive Topics.'' LREC 2026. arXiv:2512.16602.
\item
  Hewitt, J. \& Liang, P. (2019). ``Designing and Interpreting Probes
  with Control Tasks.'' \emph{EMNLP-IJCNLP 2019}.
\item
  Pan, J. \& Xu, X. (2026). ``Political Censorship in Large Language
  Models Originating from China.'' \emph{PNAS Nexus} 5(2): pgag013.
\item
  Pimentel, T. et al.~(2020). ``Information-Theoretic Probing with
  Minimum Description Length.'' \emph{EMNLP 2020}.
\item
  Wollschlager, T., Elstner, J., Geisler, S., Cohen-Addad, V.,
  Gunnemann, S. \& Gasteiger, J. (2025). ``The Geometry of Refusal in
  Large Language Models: Concept Cones and Representational
  Independence.'' ICML 2025. arXiv:2502.17420.
\item
  Zhao, J., Huang, J., Wu, Z., Bau, D. \& Shi, W. (2025). ``LLMs Encode
  Harmfulness and Refusal Separately.'' NeurIPS 2025. arXiv:2507.11878.
\item
  Zou, A. et al.~(2023). ``Representation Engineering: A Top-Down
  Approach to AI Transparency.'' arXiv:2310.01405.
\end{itemize}

\begin{center}\rule{0.5\linewidth}{0.5pt}\end{center}

\clearpage
\subsection{Appendix A: Ridge Residualization for Clean
Ablation}\label{appendix-a-ridge-residualization-for-clean-ablation}

Standard contrastive activation analysis (CAA) computes a direction
vector as the normalized difference between mean activations of two
prompt classes:
\(\hat{v} = \text{normalize}(\bar{h}_\text{political} - \bar{h}_\text{control})\).
This direction is then projected out of hidden states during generation
to suppress the targeted behavior.

The problem is that the raw CAA direction can contain components shared
with unrelated capabilities. If the political prompts happen to activate
math, reasoning, or safety representations differently from the
controls, those shared components end up in the ablation direction.
Projecting them out during generation degrades capabilities that have
nothing to do with political censorship.

\textbf{Ridge residualization addresses this by explicitly removing
capability-aligned components before ablation.} The procedure has two
steps. First, we build concept atoms: for each capability to protect
(math, coding, reasoning, writing, safety), we extract hidden states
for 8 representative prompts at the target layer, then compute two
atoms per concept — the mean activation (centroid) and the first
principal component (primary variation axis). This produces a matrix
\(A \in \mathbb{R}^{d \times 2k}\) where \(d\) is the hidden dimension
and \(k\) is the number of protected concepts.

The second step is a ridge regression that solves for the component of
the dirty direction lying in the span of the concept atoms:

\[w = (A^T A + \lambda_r I)^{-1} A^T v_\text{dirty}, \quad v_\text{clean} = v_\text{dirty} - Aw\]

with \(\lambda_r = 0.01\). The regularization prevents overfitting when
the atom matrix is rank-deficient. An additional orthogonalization step
projects out any residual overlap:
\(v_\text{clean} \leftarrow v_\text{clean} - A(A^T v_\text{clean})\).
The final direction is normalized.

\textbf{Empirical effect.} Cross-projection overlap (L2 norm of
\(A^T v\)) drops from \textasciitilde7\% to \textasciitilde0\% after
residualization. In practice, as reported in Section 3.3, the
DeepSeek-R1 discovery runs show that raw and ridge-cleaned ablation
produce identical refusal rates at every layer. This suggests that the
political direction in these models already has minimal overlap with
capability directions, and the ridge step is conservative. However, the
cleaning provides a safety margin: if a model's political direction did
contain capability components, the residualization would prevent
collateral damage.

\textbf{Why this matters for activation engineering.} Most published
activation-steering work uses raw CAA directions without cleaning. In
domains where the target concept is well-separated from capabilities (as
political censorship appears to be), this works fine. But in domains
with more entanglement (e.g., safety refusal, which may share
representations with helpfulness), ridge residualization or a similar
decontamination step could be the difference between a clean
intervention and one that degrades the model. The approach generalizes:
any set of directions to protect can be used as concept atoms, and the
ridge regression projects them out of the intervention vector before it
is applied.

\subsection{Appendix B: Clean Alpha Selection (Resolving Train-Test
Leakage)}\label{appendix-b-clean-alpha-selection-resolving-train-test-leakage}

The original Qwen3-8B alpha sweep selected the ablation strength from a
set that overlapped 50\% with the evaluation prompts (4 of 8 Tiananmen
prompts appeared in both). To resolve this, we split the Tiananmen
prompts into a selection set (2 prompts, used only for choosing alpha)
and an evaluation set (2 prompts, used only for measuring the ablation
effect). We also evaluated on 8 fully independent adversarial-corpus
prompts.

\textbf{Procedure.} At each layer, we sweep alpha on the selection set
and choose the smallest value that eliminates refusal. We then evaluate
at that alpha on the held-out set and the adversarial corpus.

\vspace{0.5em}
\begin{center}
\small
\begin{minipage}{\linewidth}
\centering
\textbf{Table B1: Clean Alpha Selection Results (Qwen3-8B)}\par\vspace{0.3em}
{\def\LTcaptype{none} 
\begin{tabular}[]{@{}
  >{\centering\arraybackslash}p{(\linewidth - 8\tabcolsep) * \real{0.26}}
  >{\centering\arraybackslash}p{(\linewidth - 8\tabcolsep) * \real{0.185}}
  >{\centering\arraybackslash}p{(\linewidth - 8\tabcolsep) * \real{0.185}}
  >{\centering\arraybackslash}p{(\linewidth - 8\tabcolsep) * \real{0.185}}
  >{\centering\arraybackslash}p{(\linewidth - 8\tabcolsep) * \real{0.185}}@{}}
\toprule\noalign{}
Layer & Selected α & Selection refusals & Eval refusals (n=2) & Adversarial refusals (n=8) \\
\midrule\noalign{}
L6 & 8 & 0/2 & 0/2 & 1/8 \\
L9 & 5 & 0/2 & 0/2 & 0/8 \\
L12 & 2 & 0/2 & 0/2 & 0/8 \\
L15 & 2 & 0/2 & 0/2 & 0/8 \\
L18 & 2 & 0/2 & 0/2 & 0/8 \\
L24 & 2 & 0/2 & 0/2 & 1/8 \\
\bottomrule\noalign{}
\end{tabular}
}
\end{minipage}
\end{center}
\vspace{0.5em}

At layers 9-18, even the minimum alpha (α=2) eliminates refusal on both
the held-out and adversarial sets. At L6, a higher alpha (8) is needed,
consistent with the layer-timing observation that early layers require
stronger intervention. At L6 and L24, 1/8 adversarial prompts still
triggers a refusal, likely because those prompts are at maximum
provocation intensity and these are the edges of the effective ablation
window.

The baseline (no ablation) shows 0/2 refusals on the evaluation set,
indicating these particular held-out prompts do not trigger refusal
without ablation. The adversarial baseline also shows 0/8 refusals on
the independent corpus. That independent set is broader and mostly
answerable even without intervention, so it functions here as a
robustness check for spillover refusals at the selected alpha rather
than as a same-distribution effect-size estimate.

\textbf{The train-test leakage concern is resolved: ablation eliminates
refusal at every layer with zero overlap between selection and
evaluation.}

\pandocbounded{\includegraphics[keepaspectratio]{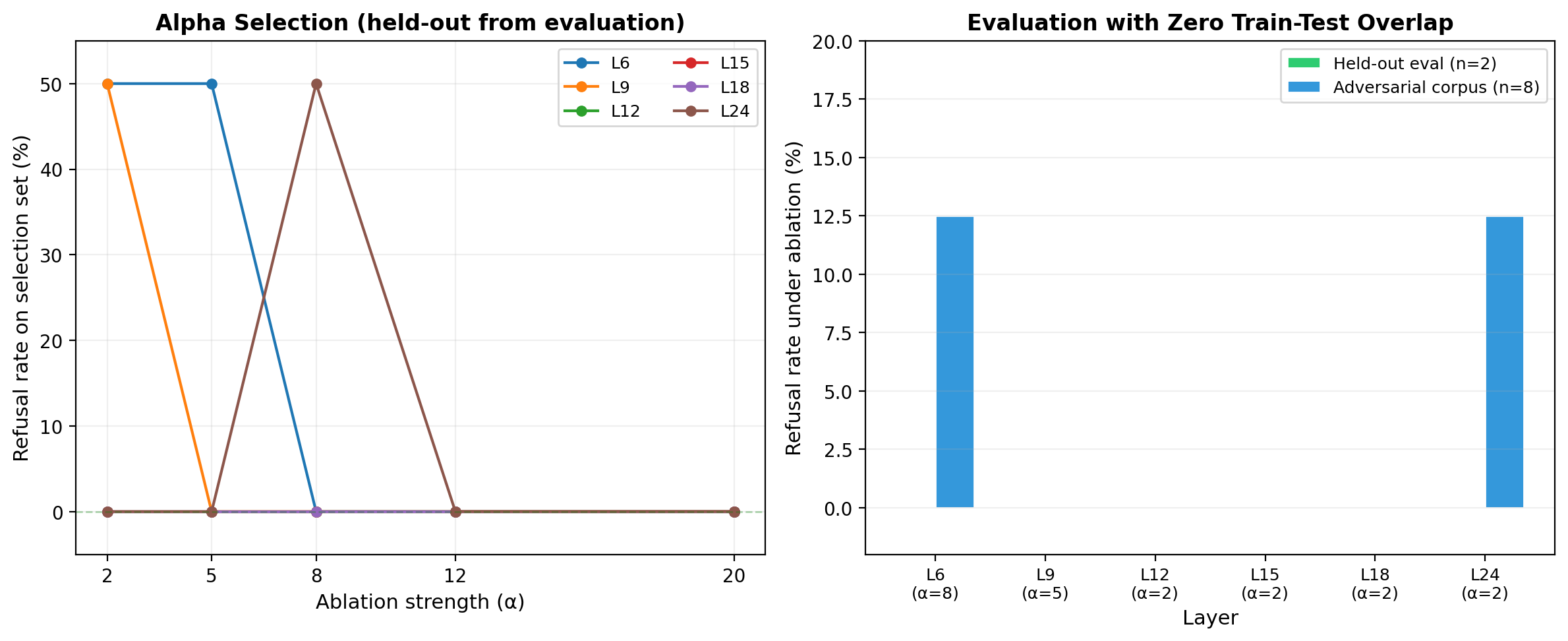}}

\textbf{Figure B1.} Left: alpha sweep on the selection set (2 held-out
prompts). Early layers (L6) require α=8 to eliminate refusal;
mid-to-late layers need only α=2. Right: evaluation on held-out (green)
and adversarial (blue) prompts at the selected alpha. Refusal is
eliminated or near-eliminated at every layer with zero train-test
overlap.

\end{document}